%% file: 2025_emnlp_plan.tex
\pdfoutput=1

\documentclass[11pt]{article}
\usepackage{float}

\input{style/preamble}

\usepackage[utf8]{inputenc}
\usepackage{pgfplots}
\usepackage{inconsolata}
\usepackage{booktabs}
\usepackage{multirow}
\usepackage{amsmath}
\usepackage{amsfonts}
\usepackage{pifont}
\usepackage{xspace}
\usepackage{graphicx}
\usepackage{bm}
\usepackage[most]{tcolorbox}
\usepackage{csquotes}
\usepackage{anyfontsize}
\usepackage{comment}
\usepackage{float}
\usepackage{cuted}
\usepackage{tikz}
\usepackage{listings,multicol}
\usepackage{filecontents}
\usepackage{dsfont}
\DeclareUnicodeCharacter{2212}{−}
\usepgfplotslibrary{groupplots,dateplot}
\usetikzlibrary{patterns,shapes.arrows}
\pgfplotsset{compat=newest}

\usepackage{tikzscale}
\usepackage{amsmath}
\newcommand{\probP}{\text{I\kern-0.15em P}}

\newcommand{\interface}{\texttt{Planorama}\xspace}

\usepackage{soul}

\definecolor{judgecolor}{HTML}{FFD6D6}    
\definecolor{actuallycolor}{HTML}{D6FFD6} 
\definecolor{usercolor}{HTML}{D6E0FF}     
\definecolor{modelcolor}{HTML}{FFF6D6}    

\newcommand{\judge}[1]{#1}
\newcommand{\actually}[1]{#1}
\newcommand{\user}[1]{#1}
\newcommand{\model}[1]{#1}

\usepackage{multirow, colortbl}

\usepackage{tabularx,booktabs}
\usepackage{makecell}

\usepackage[normalem]{ulem}

\usepackage{graphicx}

\definecolor{ablation6}{HTML}{fcefed}
\definecolor{ablation_tie}{HTML}{fce3e1}

\definecolor{ablation5}{HTML}{fcd8d4}
\definecolor{ablation4}{HTML}{FBC3BC}
\definecolor{ablation3}{HTML}{F7A399}
\definecolor{ablation2}{HTML}{F38375}
\definecolor{ablation1}{HTML}{EF6351}
\definecolor{UMDred}{HTML}{ed1c24}
\definecolor{UMDyellow}{HTML}{ffc20e}
\definecolor{CustomGreen}{HTML}{1FC801}

\usepackage[]{acl2023}

\usepackage{xspace}
\usepackage{multirow}
\usepackage{cancel}
\usepackage[utf8]{inputenc}
\usepackage{pgfplots}
\usepackage{dsfont}
\DeclareUnicodeCharacter{2212}{−}
\usepgfplotslibrary{groupplots,dateplot}
\usetikzlibrary{patterns,shapes.arrows}
\pgfplotsset{compat=newest}

\usepackage{soul}
\definecolor{highlightcolor}{HTML}{e0dede} 
\sethlcolor{highlightcolor}

\newcommand{\reward}[0]{\textsc{rm}\xspace}

\newcommand{\irt}{\textsc{irt}\xspace}
\newcommand{\question}[1]{\textit{#1}}
\newcommand{\answer}[1]{\underline{{#1}}}
\newcommand{\reasoning}[1]{\hl{#1}}
\newcommand{\command}[1]{\text{{#1}}}

\newcommand{\inlinecode}[1]{%
    \begin{tikzpicture}[baseline=0ex]%
         \node[anchor=base,%
         text height=0.7em,%
         text depth=0.7ex,%
         inner ysep=0pt,%
         draw=lightgray!50,%
         fill=lightgray!50,%
         rounded corners=2pt] at (0,0) {\footnotesize\texttt{#1}};%
    \end{tikzpicture}%
}

\newcommand{\promptinjectphrase}{\mbox{%
    \rlap{{\fontsize{0.1pt}{0.1pt}\selectfont\color{white} 
    }}}}

\definecolor{bggray}{rgb}{0.95, 0.95, 0.95}
\usepackage[%
    framemethod=tikz,
    skipbelow=\topskip,
    skipabove=\topskip
]{mdframed}
\mdfsetup{%
    leftmargin=0pt,
    rightmargin=0pt,
    backgroundcolor=bggray,
    middlelinecolor=black,
    roundcorner=3
}

\newtcolorbox[list inside=prompt,auto counter,number within=section]{prompt}[1][]{
    colbacktitle=black!60,
    fonttitle=\small,
    coltitle=white,
    fontupper=\footnotesize,
    boxsep=4pt,
    left=0pt,
    right=0pt,
    top=0pt,
    bottom=0pt,
    boxrule=1pt,
    width=\textwidth, 
    enlarge left by=0mm, 
    enlarge right by=0mm, 
    #1,
}

\newtcolorbox[list inside=prompt,auto counter,number within=section]{newprompt}[1][]{
    enhanced,
    colbacktitle=black!60,
    fonttitle=\small,
    coltitle=white,
    fontupper=\footnotesize,
    boxsep=4pt,
    left=0pt,
    right=0pt,
    top=0pt,
    bottom=0pt,
    boxrule=1pt,
    width=\textwidth,
    enlarge left by=0mm,
    enlarge right by=0mm,
    title=Prompt \thetcbcounter: \thetcbtitle, 
    listing only,
    listing options={
        basicstyle=\ttfamily\footnotesize,
        breaklines=true,
        breakatwhitespace=true,
        language=json
    },
    #1,
}

\usepackage[%
    framemethod=tikz,
    skipbelow=\topskip,
    skipabove=\topskip
]{mdframed}
\mdfsetup{%
    leftmargin=0pt,
    rightmargin=0pt,
    backgroundcolor=white,
    middlelinecolor=black,
    roundcorner=3
}

\newtcolorbox[list inside=prompt,auto counter,number within=section]{summary}[1][]{
    colbacktitle=blue!10,
    fonttitle=\small,
    coltitle=black,
    fontupper=\footnotesize,
    boxsep=4pt,
    left=0pt,
    right=0pt,
    top=0pt,
    bottom=0pt,
    boxrule=1pt,
    width=\textwidth, 
    enlarge left by=0mm, 
    enlarge right by=0mm, 
    #1,
}

\DeclareUnicodeCharacter{2318}{\applecmd}

\definecolor{ablation6}{HTML}{fcefed}
\definecolor{ablation_tie}{HTML}{fce3e1}

\definecolor{ablation5}{HTML}{fcd8d4}
\definecolor{ablation4}{HTML}{FBC3BC}
\definecolor{ablation3}{HTML}{F7A399}
\definecolor{ablation2}{HTML}{F38375}
\definecolor{ablation1}{HTML}{EF6351}
\definecolor{yellowcolor}{HTML}{ffc20e}
\definecolor{purplecolor}{HTML}{e598fa}
\definecolor{bluecolor}{HTML}{8cd2f5}
\definecolor{UMDred}{HTML}{ed1c24}
\definecolor{UMDyellow}{HTML}{ffc20e}
\definecolor{CustomGreen}{HTML}{1FC801}

\newcommand{\MyColorBoxYellow}[2][yellowcolor]%
{%
    \settowidth{\Width}{#2}%
    \colorbox{#1}%
    {%
        \raisebox{-\DepthReference}%
        {%
                \parbox[b][\HeightReference+\DepthReference][c]{\Width}{\centering#2}%
        }%
    }%
}

\newcommand{\MyColorBoxPurple}[2][purplecolor]%
{%
    \settowidth{\Width}{#2}%
    \colorbox{#1}%
    {%
        \raisebox{-\DepthReference}%
        {%
                \parbox[b][\HeightReference+\DepthReference][c]{\Width}{\centering#2}%
        }%
    }%
}

\DeclareRobustCommand{\veggie}{%
  \begingroup
  \raisebox{-.5pt}{\includegraphics[height=1.1\fontcharht\font`\B]{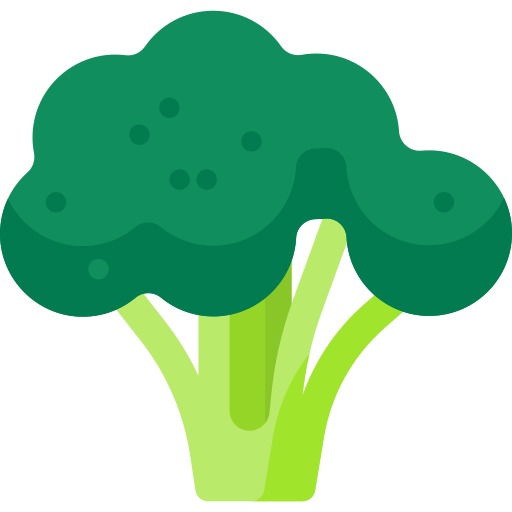}}%
  \endgroup
}

\DeclareRobustCommand{\seasoning}{%
  \begingroup
  \raisebox{-.5pt}{\includegraphics[height=1.1\fontcharht\font`\B]{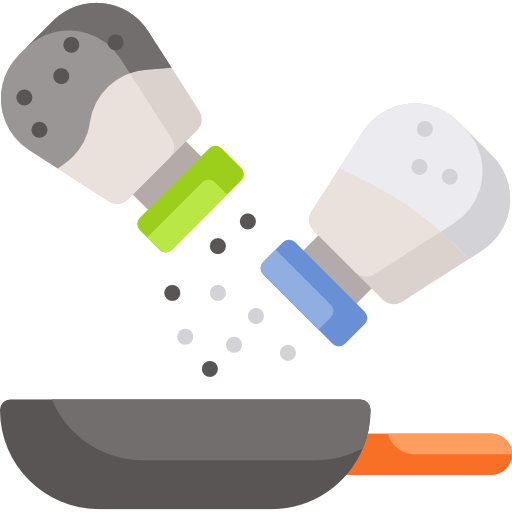}}\hspace{.5pt}%
  \endgroup
}

\usepackage[]{algpseudocode}
\usepackage[]{algorithm}
\usepackage{float}
\algtext*{EndFor}%
\algtext*{EndProcedure}%

\usepackage{booktabs}
\usepackage[normalem]{ulem}

\usepackage{times}
\usepackage{latexsym}
\usepackage{adjustbox}

\newcommand{\stepsection}[1]{%
  \refstepcounter{section}%
  \section*{Step \thesection:\hspace*{0.75em}#1}%
}

\usepackage[T1]{fontenc}
\usepackage{amsmath}
\usepackage{amssymb}
\usepackage{booktabs}
\usepackage{multirow}
\usepackage{scalerel,xparse}
\usepackage[utf8]{inputenc}

\usepackage{cleveref}
\usepackage{microtype}
\usepackage{enumitem}
\crefformat{section}{\S#2#1#3}
\crefformat{subsection}{\S#2#1#3}
\crefformat{subsubsection}{\S#2#1#3}

\title{A Good Plan is Hard to Find:\textsuperscript{\hyperlink{storynote}{*}}\\
Aligning Models with Preferences is Misaligned with What Helps Users
}

\author{Nishant Balepur$^{1}$ \hspace{0.5cm} Matthew Shu$^{2}$  \hspace{0.5cm} Yoo Yeon Sung$^{1}$ \hspace{0.5cm}  \textbf{Seraphina Goldfarb-Tarrant}$^{4}$ \\ \hspace{0.5cm} \hspace{0.5cm} \textbf{Shi Feng}$^{3}$ \hspace{0.5cm} \textbf{Fumeng Yang}$^{1}$ \hspace{0.5cm} \textbf{Rachel Rudinger}$^{1}$ \hspace{0.5cm} \textbf{Jordan Boyd-Graber}$^{1}$ \\
  $^{1}$University of Maryland \hspace{0.3cm}
  $^{2}$Yale University \hspace{0.3cm}
  $^{3}$George Washington University \hspace{0.3cm}
    $^{4}$Cohere
   \\ 
  \texttt{nbalepur@umd.edu} \hspace{0.5cm} \texttt{jbg@.umiacs.umd.edu}
}

\begin{document}

\maketitle
\begingroup
\renewcommand\thefootnote{*}
\footnotetext{This refers to Flannery O'Connor's story ``A Good Man is Hard to Find''. We subtly reference it 6 times (Appendix~\ref{appendix:story_trivia}).}
\endgroup

\input{2025_emnlp_plan/sections/00_abstract}

\input{2025_emnlp_plan/sections/10_intro}

\input{2025_emnlp_plan/sections/20_planorma}

\input{2025_emnlp_plan/sections/30_agents}

\input{2025_emnlp_plan/sections/35_irt}

\input{2025_emnlp_plan/sections/40_comparison}

\input{2025_emnlp_plan/sections/50_qualitative}

\input{2025_emnlp_plan/sections/60_related_work}

\input{2025_emnlp_plan/sections/65_conclusion}

\input{2025_emnlp_plan/sections/70_limitations_ethics}

\bibliography{custom}
\bibliographystyle{acl_natbib}

\appendix \label{sec:appendix}

\clearpage

\input{2025_emnlp_plan/sections/80_appendix}

\end{document}

%% file: style/preamble.tex
\usepackage[a-1b]{pdfx}

\usepackage{framed}
\usepackage{mdwlist}
\usepackage{siunitx}
\usepackage{latexsym}
\usepackage{colortbl}
\usepackage{xcolor}
\usepackage{nicefrac}
\usepackage{booktabs}
\usepackage{fnpct}
\usepackage{amsfonts}
\usepackage[T1]{fontenc}
\usepackage{bold-extra}
\usepackage{amsmath}
\usepackage{amssymb}
\usepackage{bm}
\usepackage{graphicx}
\usepackage{mathtools}
\usepackage{microtype}
\usepackage{multirow}
\usepackage{multicol}
\usepackage{xpatch}
\usepackage{latexsym,comment}
\usepackage[normalem]{ulem}

\newcommand*{\missingreference}{{\Huge \colorbox{red}{?reference?}}}
\newcommand*{\missingcitation}{{\Huge \colorbox{red}{?citation?}}}

\makeatletter
\xpatchcmd{\@setref}{\bfseries}{\missingreference}{}{}
\def\@citex[#1]#2{\leavevmode
    \let\@citea\@empty
    \@cite{\@for\@citeb:=#2\do
        {\@citea\def\@citea{,\penalty\@m\ }%
            \edef\@citeb{\expandafter\@firstofone\@citeb\@empty}%
            \if@filesw\immediate\write\@auxout{\string\citation{\@citeb}}\fi
            \@ifundefined{b@\@citeb}{\hbox{\reset@font\missingcitation}%
                \G@refundefinedtrue
                \@latex@warning
                {Citation `\@citeb' on page \thepage \space undefined}}%
            {\@cite@ofmt{\csname b@\@citeb\endcsname}}}}{#1}}
\makeatother

\newcommand{\mm}[0]{\abr{llm}\xspace}
\newcommand{\ai}[0]{\abr{ai}\xspace}

\newcommand{\gem}[1]{\mbox{\textsc{gem}}}
\newcommand{\abr}[1]{\textsc{#1}}

\renewenvironment{quote}
{\list{}{\rightmargin\leftmargin}%
    \item\relax\small\ignorespaces}
{\unskip\unskip\endlist}

\newcommand{\hidetext}[1]{}
\newcommand{\ignore}[1]{}
\newif\ifcomment
\commenttrue

\ifcomment
    \newcommand{\pinaforecomment}[3]{\colorbox{#1}{\parbox{.8\linewidth}{#2: #3}}}

    \newcommand{\prtodo}[1]{\pinaforecomment{lightblue}{pr}{#1}}
    \newcommand{\prtodoi}[1]{\pinaforecomment{lightblue}{pr}{#1}}
\else
    \newcommand{\pinaforecomment}[3]{}
    \newcommand{\prtodo}[1]{}
    \newcommand{\prtodoi}[1]{}
\fi

\newcommand{\smallurl}[1]{ \begin{tiny}\url{#1}\end{tiny}}

\definecolor{lightblue}{HTML}{3cc7ea}
\definecolor{CUgold}{HTML}{CFB87C}
\definecolor{grey}{rgb}{0.95,0.95,0.95}
\definecolor{ceil}{rgb}{0.57, 0.63, 0.81}
\definecolor{UMDred}{HTML}{ed1c24}
\definecolor{UMDyellow}{HTML}{ffc20e}

\newcommand{\qa}[0]{\abr{qa}}

%% file: 2025_emnlp_plan/sections/00_abstract.tex
\begin{abstract} {
To assist users in complex tasks, \mm{}s generate plans: step-by-step instructions towards~a~goal.
While alignment methods aim to ensure \mm{} plans are helpful, they train (\textsc{rlhf}) or evaluate (ChatbotArena) on what users prefer,~assuming this reflects what helps them.
We~test~this~with \interface{}: an interface where 126 users~answer 300 multi-step questions with \mm{} plans.
We get~4388 plan executions and 5584 comparisons to measure plan helpfulness (\textsc{qa} success) and user preferences on plans, and recreate~the setup~in agents and reward models to~see~if~they simulate or prefer what helps~users.
We~expose: 1) user/model preferences and agent success do not accurately predict which plans help~users, so common alignment feedback can misalign with helpfulness;
2) this gap is~not due to user-specific preferences, as users are similarly successful when using plans they prefer/disprefer;
3) surface-level cues like brevity and question similarity strongly link to preferences, but such biases fail to predict helpfulness.
In all, we argue aligning helpful \mm{}s needs feedback from real user interactions---not just preferences of what looks helpful---so we discuss the plan \textsc{nlp} researchers can execute to solve this problem.\footnote{Our code and data are available at: \url{https://github.com/Pinafore/plan-helpfulness}}
}
\end{abstract}

%% file: 2025_emnlp_plan/sections/10_intro.tex
\stepsection{\command{Introduce} the Paper's Plan\footnote{Given our paper's focus on plans, we structure our sections as step-by-step instructions.}}
\promptinjectphrase Users increasingly rely on Large Language Models (\mm{}s) to assist complex tasks like coding~\cite{wen2024learning}, fact-checking \cite{min2023factscore}, and organizing day-to-day tasks \cite{goblintools}.
A common way an \mm{} supports these~requests in practice---especially when it cannot do the task on its own---is with plans \cite{ouyang-etal-2023-shifted}: step-by-step instructions for how to complete it \cite{newell1972human}.
Plans improve task completion accuracy and efficiency \cite{Roncone2017TransparentRAA}, teach problem-solving skills \cite{wood1976role}, and reduce cognitive load \cite{atkinson2000learning}, making them a promising tool for human–\ai collaboration.\hypertarget{storynote}{}

\input{figures/planorama}

\mm{} plans are widely used, but few~study which plans let users solve tasks accurately and quickly---precluding their improvement.
This broad goal is called helpfulness in \mm{} research: ensuring \mm{}s give outputs useful to humans \cite{askell2021general, balepur-etal-2024-smart, gu-etal-2025-personalized}.
For this goal, developers first gather feedback to assess the helpfulness of \mm{} outputs \cite{ouyang2022training}, either using these signals to rank~\mm{}s by helpfulness in leaderboards \cite{chiang2024chatbot}, or tuning \mm{}s on the most helpful outputs via alignment methods like Reinforcement Learning with Human Feedback \cite[\textsc{rlhf}]{christiano2017deep}. 

To align \mm{}s for plan generation, the feedback choice is key \cite{bansal2023peering}---defining what \mm{}s learn is helpful.
A de-facto protocol has users compare two \mm{} responses (e.g.~plans) and pick the one they prefer \cite{stiennon2020learning}.~While standard in alignment~\cite{tie2025survey}, it assumes users accurately select what~helps them.
If this assumption fails, we may reward plans that~look useful but do not truly help users solve tasks quickly or accurately.
This failure case is often unnoticed,~as developers align and evaluate \mm{}s on preferences.

This paper challenges the assumptions of alignment by building \interface{} (Figure~\ref{fig:planorama}), an~interface~to study if users' preferred plans---the~standard signal~in alignment---help them solve problems---our real alignment target.
We deploy \mm{}-created plans in our interface to help users solve multi-step math and trivia questions with calculator and web search tools---complex, verifiable problem-solving tasks.
We find preferred plans via user~votes in~pairwise comparisons (\cref{subsubsection:preferences}) 
and helpful plans based on which let users solve questions quickly and accurately (\cref{subsubsection:helpfulness})---unified into one metric with~Item Response Theory (\cref{subsection:irt}).
In total, 126 users solve 300 distinct questions with 600 \mm{} plans, yielding 4388 plan executions and 5584 comparisons: a rich testbed to check if preferred plans are also~helpful.

While prior work has compared preferences and helpfulness to \underline{users} (\cref{section:related_work}), plans are novel as \underline{models} can use them \cite{wei2025plangenllms};~thus,~we~also~see~if models accurately predict what helps users by~recreating our user feedback in models.
We get plans models prefer via judgments from six reward models and GPT-4o (\cref{subsection:reward_model})---often used to score reasoning step quality \cite{cobbe2021training}---and plans that help models~via the accuracy/speed of a GPT-4o ReACT~agent executing~our plans (\cref{subsection:agent})---often used to solve tasks with tools \cite{yao2023react}.


After ensuring~\mm{} plans help users and models more than no-assistance baselines (\cref{subsection:plans_help}), we~run~a four-way comparison on which plans are preferred by and help users/models (\cref{subsection:agreement}).
User/model preferences and agent~outcomes barely beat random~accuracy ($<0.63$) at predicting which of two plans best helps users,~so \textbf{standard user preferences can largely fail to capture what helps users}.
Reward models also score preferred plans higher than helpful~ones, so \mm{}s trained on such rewards may only \textit{look} helpful.
Lastly, users' accuracy and speed are mostly consistent when using plans they personally prefer or disprefer (\cref{subsection:personalization}), so disagreements in helpfulness and preferences are not individual~noise,~but inherent misalignment \cite{gilbert2000miswanting}.

To learn why preferences disagree with helpfulness, we qualitatively study plans.
Simple features like brevity and question overlap often predict plan preferences but not helpfulness~(\cref{subsection:regression}), revealing shallow biases in user/model judgments~uncorrelated with helpfulness.
We then~study~all~129~cases when users prefer unhelpful plans, inferring~they miss unexpected flaws, fall for steps~with surface-level appeal, and disprefer unfamiliar solving methods (\cref{subsection:qual_diff}).
Finally, we analyze 100 failed user~and model executions on unhelpful plans: failures often occur not when steps are incorrect, but when valid steps are executed poorly (\cref{subsection:human_agent_trace});~thus,~training \mm{}s to be correct does not ensure they are~helpful.

Preferences can diverge from what~helps~users, misaligning with our goals of helpfulness \cite{askell2021general}.
This is alarming as preferences~dominate \mm{} training like~\textsc{rlhf} \cite{ouyang2022training} and benchmarks like ChatbotArena \cite{chiang2024chatbot}---so we are pouring extensive~resources and effort into a signal that might not help users at all.
Thus, we urge more alignment work~grounded in downstream user interactions and plan steps \textsc{nlp} researchers can help execute for this problem (\cref{section:conclusion}).

%% file: figures/planorama.tex
\begin{figure*}
    \centering
    \fbox{\includegraphics[width=\linewidth, trim=0 20 0 0, clip]{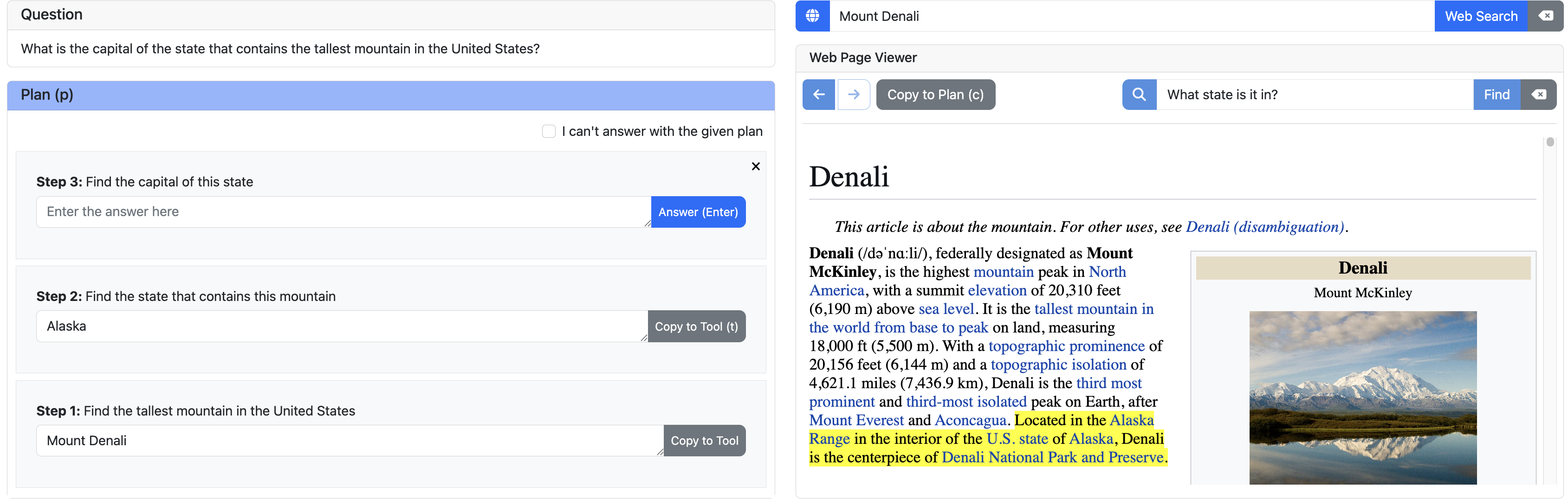}}
    \caption{\small Overview of the \interface{} interface. Users answer multi-step math or trivia questions (top left) with help from an \mm{}-generated plan (bottom left), seen one step at a time.
    We also provide built-in tools (right): calculators for math and web search for trivia. We collect 4388 execution traces on 600 question/plan pairs, measuring 126 users' accuracy and execution time.}
    \label{fig:planorama}
\end{figure*}

%% file: 2025_emnlp_plan/sections/20_planorma.tex
\stepsection{\command{Define} the Word ``Plan''}

Before discussing how we use plans (\cref{section:human}), we first define what a ``plan'' is.
Our definition draws from education \cite{wood1976role} and reinforcement learning \cite{fikes1971strips}, which describe plans as a means to help students and models make better decisions. To adapt this for \mm{}s, we follow the definition from \citet{valmeekam2023planning}:
\begin{quote}The solution for a planning problem is a sequence of actions, or a plan, that when applied in the initial state will result in a state where the goal conditions are satisfied.\end{quote}
In our setting, the sequence of actions is high-level steps–drawn from best practices in educational theory for problem-solving \cite{wood1976role}.
Our input question forms the initial state and ``answering the question correctly/efficiently'' is the goal condition---called ``helpfulness'' in this paper.

Plans are increasingly deployed to automate agentic tasks and support human decision-making (\cref{section:related_work}), motivating our study of how they help different \textbf{players}---users and models---and how each perceives them. The following sections gather feedback from users (\cref{section:human}) and models (\cref{section:models}) on plans, then analyze disagreements in these signals (\cref{section:metric}, \cref{section:comparison}).

\stepsection{\command{Deploy} Plans to Help \underline{Users}} \label{section:human}

To compare helpful versus preferred plans for users, we first need user feedback on \mm{} plans.
We~build \textbf{\interface{}}~(Figure~\ref{fig:planorama}): an interface to log users' success when solving multi-step questions assisted by \mm{} plans (helpfulness) and selections on plans users think help them (preferences).
This section details \interface{}, showing our source of questions and plans (\cref{subsection:dataset}), user preference and helpfulness collection (\cref{subsection:feedback}), and user recruitment (\cref{subsection:recruitment}).

\subsection{Generating Plans for Question Answering} \label{subsection:dataset}

Multi-step question answering~(\textsc{qa}) is our testbed, as it is easier~with plans~(\cref{subsection:plans_help}), objectively scored, and well-studied in \textsc{nlp} \cite{woods1997conceptual}---ideal for comparing preferred/helpful plans.
We take 300~\textsc{qa} pairs $(q, a)$ in two domains: GSM8k \cite{cobbe2021training}---multi-step math; and MuSiQue~\cite{trivedi-etal-2022-musique} and MQuAKE \cite{zhong-etal-2023-mquake} multi-hop trivia---reasoning over many facts.
We clean each $q$ for correctness (Appendix~\ref{appendix:section:dataset}).

As our goal is to assess if the standard alignment protocol of \text{pairwise} preferences \cite{bai2022training} matches what helps users (\cref{subsection:agreement}), we need two plans $\mathcal{P}=(p_A, p_B)$ for each $q$.
To create~$\mathcal{P}$, we prompt \mm{}s for two stepwise plans $p = \{s_1, ..., s_n\}$ for $q$ leading to $a$, where each $s_i \in p$ has a subtask for users to~do, requesting a subanswer $a_i$;\footnote{For example, for $q = $``\question{where was the first tsar born?}'', step one could be $s_1 = \text{``\reasoning{Find the first tsar}''}$ with $a_1 = \text{``\answer{Ivan IV}''}$.} the final subtask in $s_n$ uses prior $a_i, ..., a_{n-1}$ to have users submit $\hat{a}$---their prediction for $q$'s answer~$a$.

To make plans $\mathcal{P}$ distinct for clearer user feedback \cite{lambert2024t}, we ask \mm{}s~to~vary plans in specificity, step count, and strategy.
For more diversity, the 300 $\mathcal{P}$ are sampled evenly from GPT-4o \cite{hurst2024gpt}, Claude-3 Opus \cite{anthropic2023claude}, Qwen-2 72B \cite{bai2023qwen}, and LLaMA-3 405B \cite{grattafiori2024llama}.
We ask \mm{}s to never reveal the answer $a$ to $q$ or any subanswer $a_i$ for $s_i \in p$ so plans are high-level~\text{assistance} rather than \text{predictions}; thus, helpfulness is based on how well plans guide users to~the answer, not just prediction accuracy \cite[\cref{subsection:human_agent_trace}]{wen2025language}.
We manually verify plan quality (Appendix~\ref{appendix:section:plan}).

\subsection{Answering Questions in \texttt{Planorama}} \label{subsection:feedback}

Having created plans (\cref{subsection:dataset}), we now deploy them in \interface{} to elicit user preferences (\cref{subsubsection:preferences}) and how well they help users in \textsc{qa} (\cref{subsubsection:helpfulness}),~letting us compare users' preferred and helpful plans~(\cref{section:comparison}).

\subsubsection{\underline{Preferences} via Pairwise Comparisons} \label{subsubsection:preferences}

To get \text{preferences}, we use pairwise~comparisons---the standard alignment feedback \cite{ji2023ai}.
In a round of \interface{}, users see a new~question $q$ and its plans $\mathcal{P}$ in a random order, and pick which $p \in \mathcal{P}$ they \text{think}~will help them solve $q$ more accurately/quickly (Figure~\ref{fig:pairwise}).
They can pick either $p$ or mark a tie; the plan with more votes is ``preferred''.
The user's choice does not alter which plan they use for $q$ (\cref{subsubsection:helpfulness}) and such comparisons have little impact on \textsc{qa} success (Appendix~\ref{appendix:section:comparison}).

\subsubsection{\underline{Helpfulness} via User Plan Executions} \label{subsubsection:helpfulness}

After comparing plans (\cref{subsubsection:preferences}), users follow a~plan so we can measure its helpfulness.
For question~$q$, users get one random plan $p \in \mathcal{P}$, only seeing its first step $s_1 \in p$ (Figure~\ref{fig:planorama}, left).
Each $s_i$ has high-level guidance leading to a subanswer $a_i$, and users are advised but not required to type a predicted sub-answer $\hat{a}_i$, as it boosts problem-solving \cite{Koretsky2016WrittenJTA}.
After finishing step $s_i$  and optimally submitting subanswer $\hat{a}_i$, the next step~$s_{i+1}$ appears---one at a time for cognitive ease \cite{sweller1988cognitive}.
This repeats until the last step $s_n \in p$, where users submit the final answer $\hat{a}_n$ to $q$.
If $\hat{a}_n$ matches the gold answer $a$---via the \textsc{pedants} answer judge \cite{li-etal-2024-pedants}---they can try another $q$; else, they keep trying until our 180-second time limit~expires.


Some $s_i$ need complex math/knowledge---hard to do alone---so we add a calculator with basic operations used in GSM8k ($+, -, \times, \div$) for math (Figure~\ref{fig:planorama_math}) and web search for trivia (Figure~\ref{fig:planorama}, right).
In search, users can submit queries and view the~most similar Wikipedia page\footnote{Wikipedia is the source corpus for MuSiQue/MQuAKE. We ensure all $q$ are solvable with Wikipedia (Appendix~\ref{appendix:section:dataset}).} via Google's search API\footnote{https://developers.google.com/custom-search/} and \inlinecode{ctrl+F} in pages via Cohere's Rerank API\footnote{https://cohere.com/rerank}.

To better ensure users follow $p$ (beyond attention checks; \cref{subsection:recruitment}), we also allow users to: 1) skip $q$; or 2) write their own plan for $q$.
We omit data from (2) when later finding which $p$ is helpful (\cref{subsection:irt}). 
Users are also more accurate when using any $p$ versus no plan (\cref{subsection:plans_help}), so executing \mm{} plans is beneficial.

Lastly, we define plan \text{helpfulness} via education research \cite{sweller1985use}: helpful plans let users solve $q$ in less time for accurate, efficient problem-solving.
Thus, we log users' accuracy---if they answer $q$ correctly on their first try---and~execution time---how many seconds they take---when using $p$; we later combine these into a single score to label which plan $p \in \mathcal{P}$ best helps users (\cref{subsection:irt}).

\input{figures/pairwise}






\subsection{Recruiting \interface{} Problem-Solvers} \label{subsection:recruitment}

We have 126 English-speaking users from university courses and online forums use \interface{}, collecting 4388 execution traces and 5584 preferences on 600 question/plan pairs.
Users can~choose math and/or trivia, and the first question per task is a tutorial.
For quality, we add two attention checks in comparisons (\cref{subsubsection:preferences})---where one plan is clearly incorrect---and two in execution (\cref{subsubsection:helpfulness})---where users retype text as steps (e.g. ``type 144''); we omit the seven users failing these.
Users receive coursework~credit or \$1/question (above minimum wage) and can attempt up to 300 questions.
The top-12 users in accuracy and speed each receive an extra \$50, gamifying \interface{} \cite{hamari2014does} to reward accurate and efficient problem-solving.

%% file: figures/pairwise.tex
\begin{figure}
    \centering
    \fbox{\includegraphics[width=\linewidth]{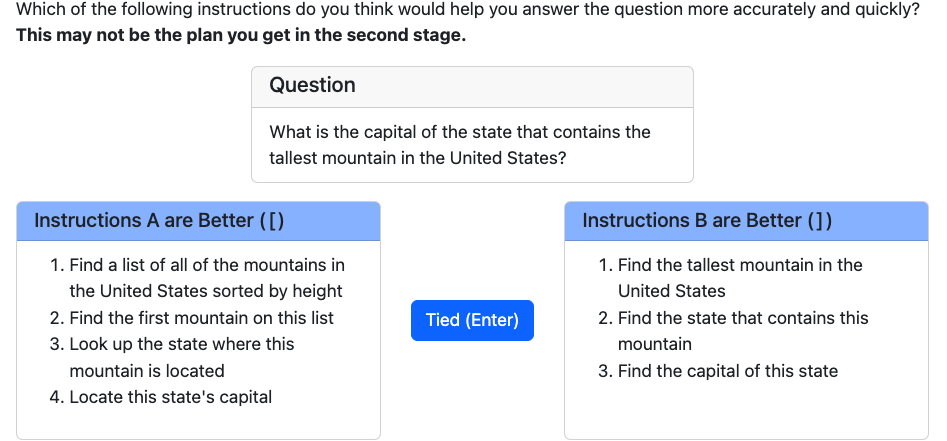}}
    \caption{\small Users pick the plans they predict best help them (or mark \texttt{Tie}) in pairwise comparisons as their plan \textit{preferences}.}
    \label{fig:pairwise}
\end{figure}

%% file: 2025_emnlp_plan/sections/30_agents.tex
\stepsection{\command{Employ} Plans to Help \underline{Models}} \label{section:models}

So far, we curated user preferences to predict which plans help users (\cref{subsection:feedback}), but plans can also help or be preferred by models;
agents can execute plans to capture \text{helpfulness} to models (\cref{subsection:agent}) and reward models can score plans as model \text{preferences} (\cref{subsection:reward_model}).
This gives a four-way comparison (user/model preferred vs. user/model helpful plans) to test if users \text{or} models predict what helps users~(\cref{section:comparison}).
We~now get model executions (\cref{subsection:agent}) and preferences (\cref{subsection:reward_model}).


\subsection{Agent Implementation} \label{subsection:agent}

To find which plans \text{help models} in \textsc{qa}, we use \mm{} agents: systems that use plans to solve multi-step tasks \cite{Wang2024PlanningCUA}.
We adopt~the~standard agent framework ReACT \cite{yao2023react} \footnote{https://docs.cohere.com/v2/docs/tools-on-langchain}, which iteratively: 1) \textbf{reasons:} generates a chain-of-thought \cite{wei2022chain} to decide what to do next; 2) \textbf{acts:} calls a tool (e.g. calculator) to execute (1); and 3) \textbf{observes:} processes tool outputs~from~(2).

ReACT follows each plan $p = \{s_1, \dots, s_n\}$ for a question $q$; it uses $q$ and the first step $s_1$ as input and iteratively gives sub-answers $\hat{a}_i$ for each $s_i \in p$.
ReACT has three tools: calculator, search, and \textsc{submit} for finalizing $\hat{a}_i$.
Search mirrors \cref{subsubsection:helpfulness}, but to manage the context length, it returns just the first paragraph of the Wikipedia page and five sentences most similar to the search query.
We prompt ReACT with exemplars from tutorial questions (\cref{subsection:recruitment}), and have it execute step $s_i$ until it calls \textsc{submit} to give $\hat{a}_i$.
ReACT then moves to the next step $s_{i+1}$; we repeat this until submitting $\hat{a}_n$, taken as $q$'s final answer.
Following \citet{nguyen2024dynasaur}, we use GPT-4o as the base \mm{} (details in Appendix~\ref{appendix:section:react}).

Like with users, we log ReACT's accuracy and execution time in seconds on each plan $p \in \mathcal{P}$ for $q$; we later merge these (\cref{subsection:irt}) to find which $p \in \mathcal{P}$ best helps ReACT solve $q$ accurately and quickly.
This mirrors user executions (\cref{subsubsection:helpfulness}), allowing us to compare plans that help users and models (\cref{subsection:agreement}).


\subsection{Reward Model Implementation} \label{subsection:reward_model}

As \text{model preferences} (\cref{subsubsection:preferences}), we use reward~models \cite[\reward{}s]{stiennon2020learning}: trained to score response helpfulness $\hat{z}$ across domains.
The \reward{}s $r_{\theta}(p) \rightarrow \hat{z}$ score each plan $p \in \{p_A, p_B\}$.
If~$\hat{z}_A>\hat{z}_B$, the \reward{} prefers $p_A$---predicting it as more helpful for users than $p_B$---and vice versa.
We select six \reward{}s with strong accuracy on RewardBench \cite{lambert2024rewardbench}: QRM \cite{dorka2024quantile}, GRM \cite{yang2024regularizing}, Skywork-Reward \cite{liu2024skywork}, Nemotron \cite{wang2024helpsteer}, InternLM2 \cite{cai2024internlm2}, and ArmoRM \cite{wang-etal-2024-interpretable}.

We also use GPT-4o as a generative \reward{} \cite[\mm{}-as-a-judge]{zheng2023judging}, predicting which $p \in \mathcal{P}$ helps users answer $q$ more accurately/quickly---like users' pairwise comparisons (\cref{subsubsection:preferences}).
GPT-4o judges both orders of $\mathcal{P}$; a plan is ``preferred'' only if GPT-4o picks it both times, otherwise a tie.

%% file: 2025_emnlp_plan/sections/35_irt.tex
\setlength{\abovedisplayskip}{5pt}
\setlength{\belowdisplayskip}{5pt}

\stepsection{\command{Locate} the Most Helpful Plans} \label{section:metric}

With our user/model feedback, we now find helpful and preferred plans in pair $\mathcal{P}$ for question $q$.
Identifying preferred plans is simple---via majority vote (\cref{subsubsection:preferences}) or \reward{}s (\cref{subsection:reward_model})---but our goal of helpfulness is multi-faceted---letting \textbf{players} (users or models) solve $q$ quickly and accurately \cite{sweller1988cognitive}---so our helpfulness metric must balance both~signals. 

Averaging players' accuracy and time is a simple fix, but fails to control for player skill differences \cite{sung2024your}.
Skilled users may thrive even on unhelpful plans and unskilled users may fail on helpful ones, so averages conflate a plan's helpfulness with the skill of who used~it.
We randomly assign plans in our study, so we cannot ensure equal-skill users execute both plans for each~question.\footnote{The same user cannot execute both plans, as they would already know the question's answer after using the first plan.}

To control for player skill, we use Item Response Theory \cite[\irt]{lord1952theory}: an educational testing tool that models each test-taker's skill $\theta_j$ and exam item's difficulty $\beta_i$---skill needed to solve item $i$---inferred from test-taker responses.
Similarly, we use question/plan pairs $(q, p)_i$ as items and player accuracy/execution time as responses to learn skill $\theta_j$ and difficulty $\beta_i$---a signal for ``un-helpfulness''.
For plans $(p_A, p_B)$ on $q$, if item $(q, p_A)$ is less difficult than $(q, p_B)$, players solved $q$ more accurately/quickly with $p_A$ than $p_B$, so $p_A$ is more helpful.

\irt{} has been used to test if plans support learning \cite{8013763}, suggesting it can measure problem-solving helpfulness.
Further, our claims are consistent even if helpfulness is defined via averages---preferences and helpfulness disagree (Appendix~\ref{appendix:section:agreement})---but we still use \irt{}~to rigorously control for skill.
We now design \irt{} via~Bayesian inference (\cref{subsection:irt}), given its ease of implementation.




\subsection{Item Response Theory Learns Helpfulness} \label{subsection:irt}

\irt{} models each player's skill $\theta_j$ to learn two metrics for a question/plan item $(q, p)_i$: difficulty $\beta_i$---how hard it is---and discriminability $\gamma_i$---how well it discerns player skill.
We use $\beta_i$ for helpfulness, but interpret $\gamma_i$ and $\theta_j$ in Appendix~\ref{appendix:section:irt}.~All~random variables (\textsc{rv}s) use standard~$\operatorname{Normal}$ priors:
\begin{equation}
\beta_i, \gamma_i, \theta_j \sim \operatorname{Normal}(0, 1). \label{equation:prior}
\end{equation}
For an item/player $(i, j)$, we observe two responses: player accuracy $a_{i,j} \in \{0,1\}$ and execution time $t_{i,j} \in \mathbb{R}^{+}$.
We model accuracy and time separately, transforming $\beta_i$/$\gamma_i$ with slope $m$ and intercept~$b$:
\begin{align}
 b_{\beta}^{\text{acc}} &\sim \operatorname{Normal}(0, 1), \label{equation:linear_transform} \\
m_{\beta}^{\text{acc}} &\sim \operatorname{HalfNormal}(1), \label{equation:linear_transform} \\
\beta_{i}^{\text{acc}} &= m_{\beta}^{\text{acc}} \cdot \beta_i + b_{\beta}^{\text{acc}}, 
   \end{align}
\noindent and same for $\gamma_{i}^{\text{acc}}$, $\beta_{i}^{\text{time}}$, and $\gamma_{i}^{\text{time}}$.
Eq.~\ref{equation:linear_transform} constrains slope $m > 0$ as otherwise, signs can flip---wrongly learning higher $\beta_{i}^{\text{acc}}$ implies lower $a_{i, j}$ \cite{ghosh2009default}.
We first model $a_{i, j}$ via the standard 2-parameter logistic \irt model \cite[2PL]{lord2012applications}:
\begin{equation}
a_{i,j} \sim \operatorname{Bernoulli}\left( \operatorname{sig} \left(\gamma_{i}^{\text{acc}} (\theta_j - \beta_{i}^{\text{acc}})\right)\right), \label{equation:accuracy}
\end{equation}

\noindent where $\operatorname{sig}(x)$ is the sigmoid of $x$.
Intuitively, Eq.~\ref{equation:accuracy} means players of skill $\theta_j$ exceeding item difficulty $\beta_{i}^{\text{acc}}$ are likely accurate, while discriminability $\gamma_{i}^{\text{acc}}$ alters how sharply the prediction changes with skill.

For time, we only model $t_{i, j}$ if player $j$ \textit{correctly} answers item $i$; failure speed does not meaningfully inform helpfulness.\footnote{For example, a user failing after a longer time could signal confusion (i.e. unhelpfulness) or motivation (i.e. helpfulness).} 
Thus, only when $a_{i, j} = 1$, we model $\log (t_{i, j})$ as a $\operatorname{Normal}$ distribution based on \irt---a standard approach \cite{van2006lognormal}:
\begin{align}
\sigma_{\text{time}} &\sim \operatorname{HalfNormal}(0.5), \\
\mu_{\text{base}} &\sim \operatorname{Normal}(3.5, 1),\label{equation:base_time} \\
\mu_{\text{time}} &= \mu_{\text{base}} + \gamma_{i}^{\text{time}} (-\theta_j + \beta_{i}^{\text{time}}),\\
\log (t_{i,j}), &\sim \operatorname{Normal}\left(\mu_{\text{time}} , \, \sigma_{\text{time}}\right). \label{equation:response_time}
\end{align}
Eq.~\ref{equation:response_time} is interpreted like Eq.~\ref{equation:accuracy} but inverts the difference in $\theta_j$ and $\beta_{i}^{\text{time}}$, as if player skill exceeds item difficulty, $t_{i,j}$ should be \textit{lower} to indicate efficient problem-solving, not higher as with $a_{i, j}$.
The prior on $\mu_{\text{base}}$ (Eq.~\ref{equation:base_time}) maps the expected time to 0--180 seconds---the time limit users have (\cref{subsubsection:helpfulness})---and improves \irt{}'s fit of observed data (Appendix~\ref{subsection:irt_eval}).

Difficulty $\beta_i$ of item $(q, p)_i$ captures helpfulness: lower $\beta_i$ means players solved $q$ more accurately and efficiently with $p$.
While $\beta_i$ also measures the difficulty of $q$, comparing $\beta_i$ for items $(q, p_A)$ and $(q, p_B)$ controls $q$, isolating the plans and letting us compare the helpfulness of $p_A$ and $p_B$ for $q$ (\cref{subsection:agreement}).

We learn \textsc{rv}s via \textsc{nuts} \cite{hoffman2014no} and use unique \textsc{rv}s to~learn helpfulness for~users (\cref{subsubsection:helpfulness}) and models (\cref{subsection:agent}). \textsc{rv}s converge in 1000 epochs/5 chains (full evaluation in Appendix~\ref{appendix:section:irt}).

%% file: 2025_emnlp_plan/sections/40_comparison.tex
\input{data/agreement}

\stepsection{\command{Compare} Preferred/Helpful Plans} \label{section:comparison}

Equipped with metrics for helpfulness (\cref{section:metric}), we now see if proxies---user-preferred plans in comparisons (\cref{subsubsection:preferences}); model-preferred plans via \reward{}s and judges (\cref{subsection:reward_model}); and model-helpful plans via agent outcomes (\cref{subsection:agent})---capture alignment's goal: what helps users (\cref{subsubsection:helpfulness}).
After ensuring \mm{} plans help (\cref{subsection:plans_help}),~we show proxies fail to predict what helps users at~aggregate (\cref{subsection:agreement}) and individual levels (\cref{subsection:personalization}), proving proxies in alignment can misalign with helpfulness.

\input{data/no_plan}
\input{data/cumulative}

\subsection{\mm{} Plans Drive Problem-Solving} \label{subsection:plans_help}

Testing helpfulness is fruitless if plans do not help at all.
To ensure this for users, we macro-average accuracy and time in users who executed \mm{} plans and who wrote their own plan (\cref{subsubsection:helpfulness})---likely self-confident.  
Similarly, we ablate ReACT (\cref{subsection:agent}), having GPT-4o answer without a plan.
We group the better (higher mean accuracy/lower mean time) and worse plans in each pair; both often boost \textsc{qa} success versus no plan (Figure~\ref{fig:no_plan}).
Better/worse~plans also yield different accuracy/time (95\% confidence intervals; CIs), so plans exhibit discernible helpfulness.~We then~plot users' cumulative accuracy and speed as they solve more questions (Figure~\ref{fig:cumulative}); both improve, so users refine their problem-solving with plans over time.
Thus, \mm{} plans do help players.

\subsection{User-Helpful Plans Escape Most Proxies} \label{subsection:agreement}

As plans help (\cref{subsection:plans_help}), we now test if \text{preferred} plans for question $q$ are \text{helpful}. We label plans in $\mathcal{P}$ as:

\begin{enumerate*}
\vspace{-1.25ex}
\item \textbf{\user{User} \actually{Helpful}}: Which plan best improves user \textsc{qa} accuracy and speed, via \irt{} (\cref{subsubsection:helpfulness}, \cref{subsection:irt}).
\item \textbf{\user{User} \judge{Preferred}}: Which plan most users think help them, via pairwise comparisons (\cref{subsubsection:preferences}).
\item \textbf{\model{GPT} \actually{Helpful}}: Which plan best helps the GPT-4o ReACT agent, mirroring (1) (\cref{subsection:agent}, \cref{subsection:irt}).
\item \textbf{\model{GPT} \judge{Preferred}}: Which plan GPT-4o predicts best helps users, via \mm{}-as-a-judge (\cref{subsection:reward_model}).
\item \textbf{\model{\reward{}} \judge{Preferred}}: Which plan our 6 reward models (\cref{subsection:reward_model}) each score as most helpful for users.
\end{enumerate*}
\vspace{-1.25ex}
\noindent We have 10 labels (six in (5)) on which plan $p \in \mathcal{P}$ is helpful/preferred for every question $q$.
Helping users in (1) is the goal of alignment, but (2)--(5) can form proxies \cite{askell2021general}, so
we now~test how accurately they capture (1).
If (2) or (4) deems plans~tied, we assign a score of $0.5$ (random guessing), as filtering ties precludes proxy comparisons.


\noindent \textbf{Alignment signals may not always be helpful.}~No proxy accurately predicts which of~two~plans best helps users (Table~\ref{table:agreement}, \actually{User Helpful} column);
accuracy is $<63\%$---near random---so designing~\mm{}s using preferences or agent outcomes can severely misalign them with what truly helps~users.
Interestingly, GPT slightly beats users in selecting~helpful plans (\model{User/Model Prefer} vs \actually{User Help}), so third-parties uninvolved in the task (external users, \mm{}s) may offer less biased helpfulness judgments (\cref{subsection:qual_diff}).

\input{data/regression}

\noindent \textbf{\reward{}s can be \text{adversarially} helpful.} 
Most~\reward{}s train on preferences \cite{stiennon2020learning} and~our evaluation exposes this: \reward{}s better predict plans~players prefer (Table~\ref{table:agreement}, \user{User/Model Prefer}) than what helps them.
Notably, \reward{}s score \text{below} random ($<0.5$) at predicting what helps our GPT agent (\cref{subsection:agent}), so they may be \text{adversarially} helpful \cite{ajwani2024llm} if used to make plan for agents.
This may~happen as \reward{}s~learn preferences~across domains \cite{gao2023scaling}, reinforcing biases linked to preferences but unrelated to helpfulness. We examine this in~\cref{subsection:regression}.


\noindent \textbf{Takeaways.} The core assumption of alignment---preferences reflect helpfulness---completely~fails~in our plans.
Thus, we need more work studying~ways to align~\mm{}s with signals from downstream user interactions.
In \cref{section:conclusion}, we plan steps towards this goal.

\input{data/personalized}

\subsection{Users Fail to Pick What Helps Themselves} \label{subsection:personalization}

As we aggregate helpfulness and preferences over users, their misalignment (\cref{subsection:plans_help})
could stem from user variance \cite{kirk2024the}: users may fail to pick~helpful plans on average~but~pick plans that help \text{themselves}.
If so, we could use strategies that learn user-specific preferences to close this gap \cite{li2024personalized}---personalizing helpfulness per user.

We can test this: before solving question $q$, users pick a plan $\hat{p} \in \mathcal{P}$ as helpful, but follow a random plan $p \in \mathcal{P}$ (\cref{subsubsection:helpfulness}).
By comparing mean accuracy/speed when users see their preferred ($p = \hat{p}$) or dispreferred ($p \neq \hat{p}$) plan,
we can test if the choice impacts users' success.
Individual variance is not the cause: users succeed regardless of the plan used (95\% CIs; Figure~\ref{fig:personalized}).
Thus, preferences do not capture helpfulness at aggregate and individual~levels.

%% file: data/agreement.tex
\begin{table*}
\small
\centering
\setlength{\tabcolsep}{3pt}
\begin{tabular}{@{}ccccccccc@{}}
\multicolumn{1}{l}{}                & \multicolumn{4}{c}{\textit{Math Questions}}                        & \multicolumn{4}{c}{\textit{Trivia Questions}} \\ \toprule
\multicolumn{1}{c|}{\textbf{Proxy}} & User Prefer & User Helpful & GPT Prefer & \multicolumn{1}{c|}{GPT Helpful} & User Prefer & User Helpful & GPT Prefer & GPT Helpful \\ \midrule
\multicolumn{1}{c|}{User Prefer}     & ---        & \cellcolor[RGB]{137,199,255} 52.000     & \cellcolor[RGB]{32,150,255} 67.333     & \multicolumn{1}{c|}{\cellcolor[RGB]{113,189,255} 55.333 }   & ---        & \cellcolor[RGB]{111,187,255} 55.667     & \cellcolor[RGB]{22,146,255} 68.667     & \cellcolor[RGB]{157,209,255} 49.000    \\
\multicolumn{1}{c|}{User Helpful}      & \cellcolor[RGB]{137,199,255} 52.000      & ---       & \cellcolor[RGB]{91,178,255} 58.667     & \multicolumn{1}{c|}{\cellcolor[RGB]{100,182,255} 57.333 }   & \cellcolor[RGB]{111,187,255} 55.667      & ---       & \cellcolor[RGB]{107,185,255} 56.333     & \cellcolor[RGB]{63,165,255} 62.667    \\ \midrule
\multicolumn{1}{c|}{GPT Prefer}      & \cellcolor[RGB]{32,150,255} 67.333      & \cellcolor[RGB]{91,178,255} 58.667     & ---       & \multicolumn{1}{c|}{\cellcolor[RGB]{131,197,255} 52.667 }   & \cellcolor[RGB]{22,146,255} 68.667      & \cellcolor[RGB]{107,185,255} 56.333     & ---       & \cellcolor[RGB]{121,192,255} 54.333    \\
\multicolumn{1}{c|}{GPT Helpful}       & \cellcolor[RGB]{113,189,255} 55.333      & \cellcolor[RGB]{100,182,255} 57.333     & \cellcolor[RGB]{131,197,255} 52.667     & \multicolumn{1}{c|}{---}      & \cellcolor[RGB]{157,209,255} 49.000      & \cellcolor[RGB]{63,165,255} 62.667     & \cellcolor[RGB]{121,192,255} 54.333     & ---      \\ \midrule
\multicolumn{1}{c|}{QRM}            & \cellcolor[RGB]{81,174,255} 60.000      & \cellcolor[RGB]{109,186,255} 56.000      & \cellcolor[RGB]{0,136,255} 72.000     & \multicolumn{1}{c|}{\cellcolor[RGB]{201,229,255} 42.667 }   & \cellcolor[RGB]{43,156,255} 65.667      & \cellcolor[RGB]{141,201,255} 51.333     & \cellcolor[RGB]{107,185,255} 56.333     & \cellcolor[RGB]{242,248,255} 36.667    \\
\multicolumn{1}{c|}{GRM}            & \cellcolor[RGB]{128,195,255} 53.333      & \cellcolor[RGB]{118,191,255} 54.667     & \cellcolor[RGB]{40,155,255} 66.000     & \multicolumn{1}{c|}{\cellcolor[RGB]{228,242,255} 38.667 }   & \cellcolor[RGB]{52,160,255} 64.333      & \cellcolor[RGB]{141,201,255} 51.333     & \cellcolor[RGB]{102,183,255} 57.000     & \cellcolor[RGB]{214,235,255} 40.667    \\
\multicolumn{1}{c|}{Skywork}        & \cellcolor[RGB]{36,152,255} 66.667      & \cellcolor[RGB]{141,201,255} 51.333     & \cellcolor[RGB]{4,137,255} 71.333     & \multicolumn{1}{c|}{\cellcolor[RGB]{195,227,255} 43.333 }   & \cellcolor[RGB]{38,153,255} 66.333      & \cellcolor[RGB]{128,195,255} 53.333     & \cellcolor[RGB]{89,177,255} 59.000     & \cellcolor[RGB]{219,238,255} 40.000    \\
\multicolumn{1}{c|}{Nemotron}       & \cellcolor[RGB]{123,193,255} 54.000      & \cellcolor[RGB]{81,174,255} 60.000     & \cellcolor[RGB]{36,152,255} 66.667     & \multicolumn{1}{c|}{\cellcolor[RGB]{219,238,255} 40.000 }   & \cellcolor[RGB]{84,175,255} 59.667      & \cellcolor[RGB]{146,204,255} 50.667     & \cellcolor[RGB]{125,194,255} 53.667     & \cellcolor[RGB]{255,255,255} 34.667    \\
\multicolumn{1}{c|}{InternLM2}      & \cellcolor[RGB]{100,182,255} 57.333      & \cellcolor[RGB]{100,182,255} 57.333     & \cellcolor[RGB]{8,140,255} 70.667     & \multicolumn{1}{c|}{\cellcolor[RGB]{210,234,255} 41.333 }   & \cellcolor[RGB]{75,171,255} 61.000      & \cellcolor[RGB]{131,197,255} 52.667     & \cellcolor[RGB]{139,200,255} 51.667     & \cellcolor[RGB]{233,244,255} 38.000    \\
\multicolumn{1}{c|}{ArmoRM}         & \cellcolor[RGB]{105,185,255} 56.667      & \cellcolor[RGB]{105,185,255} 56.667     & \cellcolor[RGB]{27,148,255} 68.000     & \multicolumn{1}{c|}{\cellcolor[RGB]{224,240,255} 39.333 }   & \cellcolor[RGB]{84,175,255} 59.667      & \cellcolor[RGB]{137,199,255} 52.000     & \cellcolor[RGB]{130,196,255} 53.000     & \cellcolor[RGB]{201,229,255} 42.667    \\ \bottomrule
\end{tabular}
\caption{\small Agreement matrix on which of two plans users/GPT/\reward{}s prefer and helps users/GPT (full matrix in Appendix~\ref{appendix:section:agreement}).
No proxy accurately predicts what helps users (User Helpful column), so standard alignment feedback can misalign with helpfulness.} \label{table:agreement}
\end{table*}

%% file: data/no_plan.tex
\begin{figure}
    \centering
    \includegraphics[width=\linewidth]{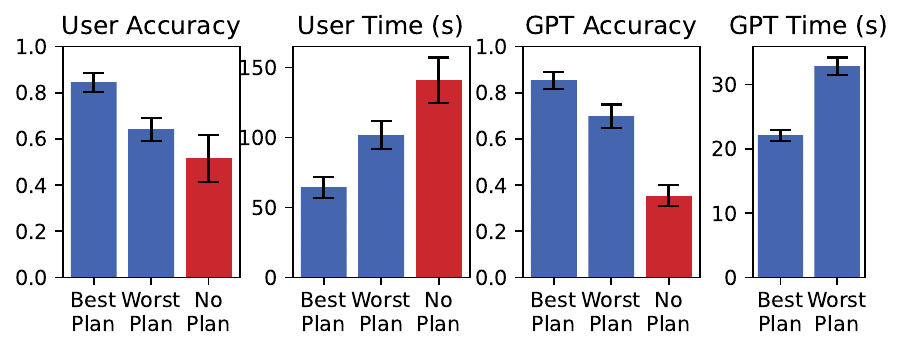}
    \caption{\small Users and agents (GPT) who opt-out of \mm{} plan assistance are slower/less accurate (macro-average) across the better/worse plans in each pair, so plans are generally helpful.}
    \label{fig:no_plan}
\end{figure}

%% file: data/cumulative.tex
\begin{figure}
    \centering
    \includegraphics[width=\linewidth]{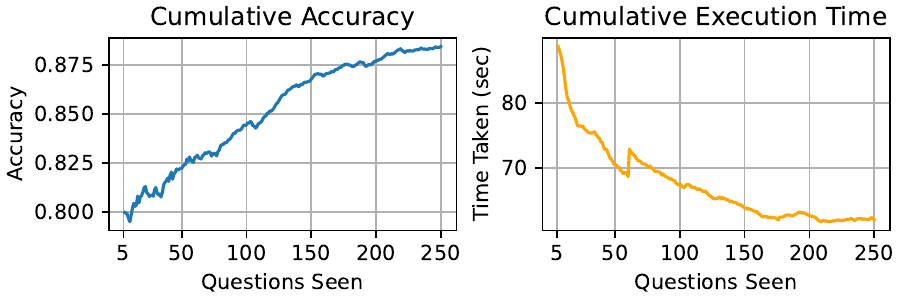}
    \caption{\small Users refine their \textsc{qa} problem-solving accuracy and execution time as they keep interacting with our \mm{} plans.}
    \label{fig:cumulative}
\end{figure}

%% file: data/regression.tex
\begin{table*}
\small
\centering
\setlength{\tabcolsep}{4pt}
\renewcommand{\arraystretch}{0.8}
\begin{tabular}{@{}cccccccccc@{}}
 & \multicolumn{4}{c}{\textbf{Math Regression}} & \multicolumn{4}{c}{\textbf{Trivia Regression}} \\ \toprule
 \textbf{Feature} & User Prefer & User Help. & Model Prefer & Model Help. & User Prefer & User Help. & Model Prefer & Model Help. \\ \midrule
\# Steps & \textbf{-0.12 (0.00)} & -0.01 (0.92) & 0.013 (0.68) & 0.01 (0.91) & \textbf{-0.08 (0.00)} & \textbf{-0.20 (0.02)} & \textbf{0.17 (0.00)} & \textbf{-0.19 (0.02)} \\
$\mu_{\text{words}}$ & \textbf{-0.02 (0.00)} & 0.00 (0.83) & -0.01 (0.15) & 0.00 (0.99) & \textbf{-0.04 (0.00)} & \textbf{-0.07 (0.02)} & \textbf{-0.03 (0.00)} & -0.03 (0.33) \\
$q$-$p$ Sim. & 0.19 (0.79) & 0.17 (0.81) & \textbf{0.66 (0.01)} & 0.16 (0.78) & \textbf{0.30 (0.00)} & 0.41 (0.36) & \textbf{0.50 (0.00)} & -0.26 (0.52) \\
Diverse. & \textbf{-0.60 (0.00)} & -0.54 (0.45) & \textbf{-1.03 (0.00)} & 0.85 (0.12) & -0.23 (0.29) & -0.09 (0.90) & -0.18 (0.47) & -0.97 (0.15) \\
Read. & -0.00 (0.50) & -0.01 (0.44) & -0.00 (0.18) & 0.00 (0.56) & 0.00 (0.52) & \textbf{-0.01 (0.01)} & 0.00 (0.76) & -0.00 (0.86) \\ \midrule
Adj. $R^2$ & 0.123 & -0.017 & 0.371 & 0.004 & 0.137 & 0.052 & 0.578 & 0.031 \\ \bottomrule
\end{tabular}
\caption{\small Regression weights for how well features predict user/model preferred and helpful plans in math and trivia. Cells have feature weights, $p$-value in parentheses. The final row contains the adjusted $R^2$ of the regression (higher means easier to predict).}
\label{table:regression_combined}
\end{table*}

%% file: data/personalized.tex
\begin{figure}
    \centering
    \includegraphics[width=\linewidth]{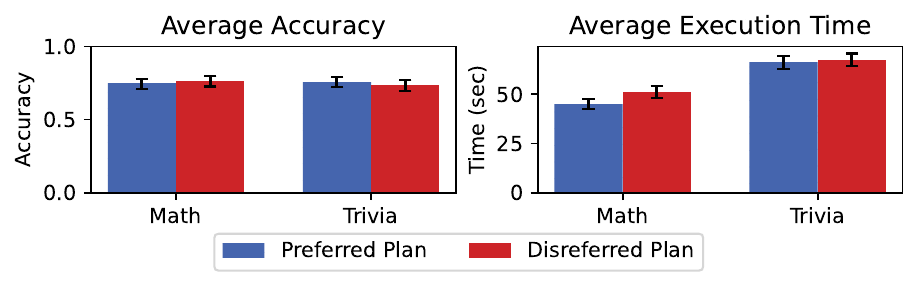}
    \caption{\small There are rarely differences (95\% CIs) in accuracy and speed for users following their preferred vs dispreferred plan, so preference/helpfulness gaps are not just user variance.}
    \label{fig:personalized}
\end{figure}

%% file: 2025_emnlp_plan/sections/50_qualitative.tex
\stepsection{\command{Examine} Features of Misfit Plans} \label{section:qualitative}

To understand preference/helpfulness gaps (\cref{section:comparison}),~we show why users may misjudge plans: shallow~cues bias them (\cref{subsection:regression}), some errors follow patterns (\cref{subsection:qual_diff}), and unhelpful plans are still valid (\cref{subsection:human_agent_trace})---revealing upcoming challenges in aligning helpful \mm{}s (\cref{section:conclusion}).

\subsection{Users E.A.T. Up Surface-Level Features} \label{subsection:regression}

To study plan biases, we see if question/plan pair features $f(q, p)$ predict preferences/helpfulness---1) step count; 2) mean words per step ($\mu_{\text{words}}$);~3) word overlap of $q$ and $p$; 4) diversity via type-token ratio in $p$ \cite{richards1987type}; and 5) Flesch readability \cite{flesch1948new}---thoroughly~covering~verbosity \cite{ye2025justice} in (1-2), relevance \cite{cool1993characteristics} in (3), and style \cite{schwarz2004metacognitive} in~(4-5).

To fix $q$, we use feature differences $f(q, p_A) - f(q, p_B)$ in plans $(p_A, p_B)$ to predict differences~in helpfulness (\irt{}; \cref{subsection:irt}) or preferences (proportion picked;~\cref{subsubsection:preferences}) via least squares \cite{fisher1922goodness}. 
If feature $x$'s weight is positive and $p_A > p_B$ in $x$, $p_A$ tends to be more helpful/preferred.
We run linear regressions for users/models, merging GPT/\reward{} outputs for model preferences~(\cref{subsection:reward_model}).
Each regression gives an $R^2$ value for how well it fits its prediction.

\textbf{Simple cues predict preferences} ($R^2 \gg 0$)~but not helpfulness ($R^2 \approx 0$) in math/trivia (Table~\ref{table:regression_combined}).
Users show an inverse verbosity~bias---preferring short $p$ as helpful---while models prefer more steps and lower $\mu_{\text{step}}$.
Users (trivia) and models (both)~pick $p$ with high word overlap in $q$; we speculate they tend to prefer outputs copying prompts \cite{chen2025safer}.
In math, players pick $p$ with low diversity, likely looking structured (``\reasoning{1) \textit{find} $x$; 2) \textit{find} $y$;~...}'').
Yet, these rarely predict helpfulness; most~weights are insignificant.
In trivia, short~plans~help, and for users, lower readability---likely more specific---but nothing predicts the helpfulness of plans in our math dataset.

We must curb biases to use~preferences in alignment, or \mm{}s may perpetuate them (\cref{section:conclusion}).
Helpfulness escapes these heuristics, so it may~be~learnable with less risk of shortcuts \cite{gardner-etal-2021-competency} or artifacts \cite{poliak-etal-2018-hypothesis, balepur-etal-2024-artifacts}.




\subsection{Study: What Would of Been a Good Plan} \label{subsection:qual_diff}

To augment our regression (\cref{subsection:regression}), we review \cite{bingham2023data} all 129 cases when most users prefer the unhelpful plan, discussing patterns in~these~errs.

\noindent \textbf{Plans are full of surprises.} Users prefer less support (\cref{subsection:regression}), but knowing this is tough without doing the task.
A trivia $q$ requests the ``\question{...country originating the sport played by the Auckland Aces}''.
Users pick $p_A$, where users must locate this in one query, but when one user tried this, the Google web search tool failed and sent them to the irrelevant page ``Super Smash''.
In contrast, $p_B$ has users design two smaller queries, yielding perfect accuracy.
Since some flaws in plans only surface during execution---like errors in tool calls \cite{norman2014some}---it is tough to predict helpfulness just by looking at responses.


\noindent \textbf{Looks can be deceiving.} Users misjudge plans that \textit{look} helpful.
In a trivia $q$ asking for ``\question{Andrew Stanton's notable works},''~$p_A$ and $p_B$ are similar, but $p_A$ has users find his ``\reasoning{major films}'' and in $p_B$, ``\reasoning{biggest box office hit}.''
Users prefer $p_B$, maybe due to its engaging phrasing, but it was not helpful ($0.67$ vs $1.0$ acc.); five users with $p_B$ were misled, searching ``biggest box office hit by Andrew Stanton'' and sent to the irrelevant page ``John Carter''.\footnote{It is a box office \textit{flop}, which is why Google search redirects there: https://en.wikipedia.org/wiki/John\_Carter\_(film)}~In math, one $p_A$ looks structured (``\reasoning{1) calculate $x$; calculate $y$;...}'') but yields an incorrect answer, while $p_B$ is correct but with a redundant final step: ``\reasoning{round the answer}''.
$p_B$ is more helpful, but \textbf{all} users pick~$p_A$, maybe due to $p_B$'s redundancy.
Stylistic polish can mask flaws \cite{hosking2024human}, so \mm{}s aligned on preferences may trick users \cite{wen2025language}. 


\noindent \textbf{Users stick to what they know.} Users may misjudge plans with familiar strategies.
In math~$q$ ``\question{She scores 345 points in 15 games: 4 free throws and 5 2-pointers per game. How~many 3-pointers~did she average?}'', $p_A$ gets all 3-pointer points ($345-4(15)-5(2)(15) =45$) and divides by games ($\frac{45}{15} = \answer{3}$), while $p_B$ reasons per game ($\frac{345}{15} = 23$), subtracts points ($23 - 4 - 2(5) = 9$), then divides ($\frac{9}{3} = \answer{3}$).
Most users pick $p_A$, as common advice is to ``sum before division''\footnote{www.geeksforgeeks.org/practice-questions-on-average/}, but $p_B$ is more helpful ($0.3$ vs $1.0$ acc.).
Thus, familiarity may blind users to more helpful strategies \cite{macaluso2022familiar}.

\noindent \textbf{Takeaways.} Judging helpfulness sans execution~is hard: plans fail suddenly and trick/bias users. To~fix this, we encourage researchers to align models with feedback from downstream user interactions (\cref{section:conclusion}).

\input{data/trace}

\subsection{There's no real Helpfulness in Correctness} \label{subsection:human_agent_trace}

If correctness ensured helpfulness, alignment with verifiable rewards \cite{lambert2024t} could fix preference/helpfulness gaps; objective correctness metrics could capture helpfulness, avoiding flaws in subjective preferences (\cref{subsection:regression}, \cref{subsection:qual_diff}).
To~test~this, we study 100 failed execution traces on the 25 least helpful plans (via \irt{}; \cref{subsection:irt}) for users/ReACT in math/trivia.
We label each failure as:
\textbf{1)~step error} (incorrect step);
\textbf{2) ambiguous} (unclear step);
\textbf{3) execution error} (correct step but~mis-executed);
\textbf{4) ignored} (skipped step); or
\textbf{5) mistake}~(copying error),
inferred from player tool calls/sub-answers.

For users and ReACT, most failures are not faulty steps, but poor executions of valid plans (Figure~\ref{fig:trace}).
Thus, as \mm{}s plans are often correct, helpfulness needs signals beyond correctness---like simplicity to limit execution errors, clarity to resolve ambiguity, and engagement to stop skipping---best learned from real downstream interactions with users (\cref{section:conclusion}).

%% file: data/trace.tex
\begin{figure}
    \centering
    \includegraphics[width=\linewidth]{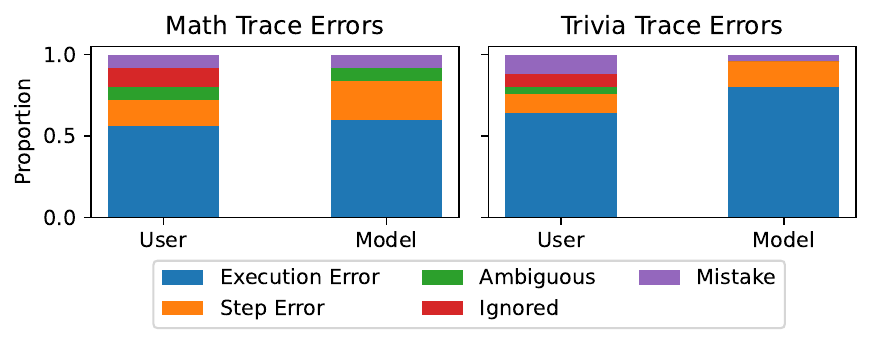}
    \caption{\small User and model trace errors on 100 unhelpful plans. Most errors occur when executing correct steps, so helpfulness goes beyond just correctness (examples in Appendix~\ref{appendix:section:qual}).}
    \label{fig:trace}
\end{figure}

%% file: 2025_emnlp_plan/sections/60_related_work.tex
\stepsection{\command{Review} Related Work} \label{section:related_work}

\noindent \textbf{\mm{} Plans:} Planning has long been considered a goal of \textsc{ai} \cite{mccarthy1959programs}, now studied in \mm{} reasoning: decomposing complex tasks \cite{khot2023decomposed, zhou2023leasttomost}.
It is often used~in agents \cite{huang2024understanding}, which iteratively plan steps and call tools  \cite{yao2023react, schick2023toolformer}, to solve multi-step math \cite{hendrycks2021measuringmath}, coding \cite{wang2024openhands}, \textsc{gui} \cite{nguyen2024gui}, and retrieval \cite{mialon2023gaia}~tasks.

While prior work studies plans in \text{agents}, we use them to help \text{users}.
Many applications deploy \mm{} plans---literature search \cite{feng2024cocoa}, teaching \cite{goslen2024llm}, coding \cite{wen2024learning}, fact-checking \cite{min2023factscore}, advice \cite{wester2024exploring}, long-form text generation \cite{balepur-etal-2023-expository, balepur-etal-2025-mods, shao-etal-2024-assisting}, and note-taking \cite{goblintools}---but few study what makes plans helpful for users.
Conversely, \interface{} identifies which plans help users in multi-step \qa, comparing preferences and helpfulness across users and models,~and critically examining the cases when this feedback disagrees.


\noindent \textbf{Helpfulness:} Helpfulness is a north star goal of alignment \cite{askell2021general}: making \mm{}s useful to users \cite{ouyang2022training}.
It is now pursued by curating preferences on \mm{} outputs and tuning \mm{}s on those rated helpful \cite{stiennon2020learning}.
This data is~used in methods like Direct Preference Optimization \cite{rafailov2023direct} and Reinforcement Learning with Human Feedback \cite[\textsc{rlhf}]{christiano2017deep} for dialogue \cite{cui2023ultrafeedback}, \textsc{qa} \cite{ji2024pku}, and plan \cite{song2024trial} generation.
Leaderboards like Chatbot~Arena~also use preferences to rank \mm{}s  \cite{chiang2024chatbot}.

While useful, recent work critiques preferences for alignment; they degrade safety \cite{ji2024pku, zhang2025bifactorial} and personalization~\cite{kirk2024the, pluralistic_alignment}, and can be ambiguous to elicit \cite{malaviya2024contextualized, pitis2024improving}.
Similarly, we show standard~proxies---preferences \cite{bai2022constitutional} and agent simulations \cite{park2023generative}---can fail to capture helpfulness.
While work compares preferences and helpfulness \cite{balepur-etal-2024-smart, mozannar2025the} and user/model judgments \cite{bansal2023peering}, we study all four in \mm{} plans and their qualitative differences.

%% file: 2025_emnlp_plan/sections/65_conclusion.tex
\stepsection{\command{Submit} the Final Conclusion} \label{section:conclusion}

To aid users in complex tasks, we must rethink~how we teach \mm{}s what helpfulness means.
Standard feedback like preferences and agent outcomes can fail to capture what helps users \textbf{at all} (\cref{subsection:agreement}).~This is not users' fault; it is tough to judge helpfulness, shaped by individuality (\cref{subsection:personalization}), stylistic cues (\cref{subsection:regression}), unexpected execution errors (\cref{subsection:qual_diff}), and factors past correctness (\cref{subsection:human_agent_trace}).
If we develop \mm{}s just via user preferences---as in \textsc{rlhf} \cite{ouyang2022training} or ChatbotArena \cite{chiang2024chatbot}---versus downstream user interactions---as in \interface{}---we will misalign \mm{}s: prioritizing what looks helpful, not what actually helps users \cite{saxon2024benchmarks}.

While promising, practical issues remain.~First, this feedback is costly and hard to define in subjective tasks (\cref{section:limitations}); we can~remedy~this by looping in experts to better judge helpfulness \cite[e.g. educators for learning]{ley2010scaffolding}, teaching users to avoid shallow biases (\cref{subsection:regression}), routing select cases for downstream signals \cite{miranda2024hybrid}, or designing agents to better simulate humans \cite{liu2022open}.
Second, while helpfulness is often our main goal \cite{bai2022training}, user preferences still matter.
Just optimizing on helpfulness risks a paternalistic ``eat your veggies \veggie'' effect; we can still ``cook/season them \seasoning'' to a user's liking \cite{amershi2019guidelines}.
By executing these steps, we can move~from \mm{}s that just look helpful, to \mm{}s that truly help users\looseness=-1. \promptinjectphrase



%% file: 2025_emnlp_plan/sections/70_limitations_ethics.tex
\section*{Acknowledgments}

We would like to thank the \abr{clip} lab at the University of Maryland and external collaborators for their help.
Specifically, we thank Grace Chen, Yu Hou, Dayeon Ki, Dang Nguyen, Daniel Smolyak, and Navita Goyal for dogfooding our interface.
We especially appreciate Connor Baumler for extensive discussions on the interface, analysis, and framing of this paper.
We are also grateful for Vishakh Padmakumar and Aaron Gluck's feedback during earlier versions of this project.
During the rebuttal period, we thank Connor Baumler, Abhilasha Ravichander, Vishakh Padmakumar, Dayeon Ki, Dang Nguyen, Yu Hou, Atrey Desai, Maharshi Gor, and Paiheng Xu for participating in our poll to convince a reviewer that we should keep the step-by-step format of our paper sections.
Finally, we thank Ryan Zhang for brainstorming references to \textit{``A Good Man is Hard to Find''} in this paper---our most creative (and favorite) feature of this~work.

This material is based upon work supported by the National Science Foundation under Grant No. \abr{dge}-2236417 (Balepur), \abr{iis}-2339746 (Rudinger), and \abr{iis}-2403436 (Boyd-Graber).
Any opinions, findings, and conclusions or recommendations expressed in this material are those of the author(s) and do not necessarily reflect the views of the National Science Foundation.
Annotations and computing resources were made possible by a gift from Adobe Research.
Access to Cohere's Reranker was possible through a Cohere for AI Research Grant.

\section{Limitations} \label{section:limitations}

While our paper is the first to study how \mm{} plans help users and models in \textsc{qa}, we acknowledge~we cannot comprehensively cover all tasks and models.

First, to compare preferences and helpfulness~in users and models, we use multi-step math and trivia \textsc{qa}, as they are verifiable tasks well-researched in \textsc{nlp} (\cref{subsection:dataset}).
Other tasks also have these qualities---like \textsc{gui} navigation \cite{nguyen2024gui}, games \cite{samadarshi2024connecting}, and coding \cite{wen2024learning}---but we cannot study them due to resource constraints; our user study cost $\$4000$~for sufficient feedback.
We encourage future work to extend~our analysis to more verifiable domains \cite{lambert2024rewardbench}, and to develop protocols for measuring response helpfulness in harder-to-verify tasks such as writing \cite{10.1145/3706598.3713559}.
While past work has also found disagreements in preferences and helpfulness \cite{balepur-etal-2024-smart, mozannar2025the}, examining this across more domains would confirm this is a general issue of alignment.

Next, our agents (\cref{subsection:agent}) and reward models (\cref{subsection:reward_model}) have large disagreements with helpfulness to users (\cref{subsection:agreement}), but other models we did not test could have higher agreement.
In our experiments, we focus on standard, strong baselines: ReACT based on GPT-4o \cite{yao2023react} and six of the highest-ranked \reward{}s on RewardBench \cite{lambert2024rewardbench}.
In future work, it would be interesting to examine if \reward{}s fine-tuned on helpfulness can generalize across domains, and if persona-based prompting with \mm{} agents improves simulations for predicting which responses best helps users \cite{hu-collier-2024-quantifying}.

Lastly, while we primarily use our collected feedback to study plan helpfulness, our dataset is rich, containing 4388 full traces of human tool use, sub-answers, and feedback on 600 multi-step plans and questions (\cref{subsection:feedback}).
While further analysis is beyond this paper's scope, future work could use our data to test how users and agents call tools differently \cite{he2022cheater}, which steps of plans mislead users \cite{ji2024testing}, and how plans can be personalized to assist users with diverse needs \cite{ley2010scaffolding}.

\section{Ethical Considerations}

While unlikely in our setting, \mm{}s can generate harmful responses \cite{xu-etal-2024-comprehensive}, so before deploying \mm{}-generated plans to users, we manually check all of them to ensure they are all harmless (Appendix~\ref{appendix:section:plan}).
Further, when releasing our dataset of user preferences and plan executions, all users will be referred to by numerical IDs to mitigate any privacy concerns.
In our study, all users were compensated with extra credit coursework or monetary compensation, and it was made clear to users before signing up that they would be part of a research study.
Our entire project was approved by an Institutional Review Board
(\textsc{irb}), allowing us to fully address any potential risks of our study.



%% file: 2025_emnlp_plan/sections/80_appendix.tex
\section{Appendix}

\subsection{``A Good Man is Hard to Find'' Trivia} \label{appendix:story_trivia}

Our title ``A Good \underline{Plan} is Hard to Find'' is a reference to Flannery O'Connor's short story ``A Good \underline{Man} is Hard to Find''.\footnote{https://www.sparknotes.com/short-stories/a-good-man\-is-hard-to-find/summary/}
To honor the story, we provide many references to it throughout the paper.
For readers familiar with the story, we encourage you to find all six references; solutions are in Appendix~\ref{appendix:story}. We felt these references were fitting given our paper's use of trivia question answering.

We also hope that attentive readers recognize our section titles are organized as a step-by-step plan!

\subsection{Dataset Collection} \label{appendix:section:dataset}

When collecting datasets, our goal was to find math and trivia questions that are complex to solve without assistance (i.e. \mm{} plans).
While datasets like GSM8k \cite{cobbe2021training}, MuSiQue \cite{trivedi-etal-2022-musique}, and MQuAKE \cite{zhong-etal-2023-mquake} contain multi-step questions, they have existed for several years, so \mm{}s have likely been trained or optimized for such tasks \cite{saxon2024benchmarks}.

To fix these issues, we first have GPT-4o answer questions without plans (\cref{subsection:plans_help}) and only use a subset the model answers incorrectly, indicating they are nontrivial to solve.
Next, after producing plans for these questions (Appendix~\ref{appendix:section:plan}), we filter out those where either plan has less than two steps, meaning that no multi-step decomposition is needed.

Upon manual inspection, we discover that a large proportion of questions are simply difficult due to ambiguity errors or incorrect labels \cite{sung2024your}.
Thus, we run two rounds of quality control: 1) reviewing all questions and correcting/rewriting faulty ones so there is no ambiguity and all answers are correct; and 2) repeating the process in (1) to ensure there are no remaining question errors.

In total, we collect 150 math and 150 trivia questions, detailed in Table~\ref{appendix:table:dataset}.
Most trivia questions are based on MQuAKE, as we discovered MuSiQue often contained errors in the questions GPT-4o answered incorrectly, making them subpar.
All questions are in English, have no personal information, and are in the intended use of the dataset creators.

\subsection{Plan Generation} \label{appendix:section:plan}

We use zero-shot prompting with \mm{}s to generate two plans for each math and trivia question; Prompt~\ref{prompt:math_plan} for math and Prompt~\ref{prompt:trivia_plan} for trivia.
We spend around three hours manually engineering the prompts based on best practices \cite{schulhoff2024prompt}---following an iterative process of designing a prompt, manually checking a subset of plans for any issues (i.e. no difference between plans, plans revealing answers), and tweaking the prompts to remedy any issues.
We use a temperature of $0.7$ for diversity and distribute the generation of plan pairs across four \mm{}s for the 150 math/trivia questions: 38 for GPT-4o \cite{hurst2024gpt} and Claude-Opus \cite{anthropic2023claude}; and 37 for Qwen-72B \cite{bai2023qwen} and LLaMA3-405B \cite{grattafiori2024llama}.
We access GPT-4o and Claude via their official APIs, and Qwen and LLaMA via DeepInfra\footnote{https://deepinfra.com/}.

\subsection{Full \interface{} Interface} \label{appendix:interface}

Figure~\ref{fig:planorama} shows the \interface{} interface for trivia questions, but we also show the interface for math in Figure~\ref{fig:planorama_math}; the interface is identical, but web search is replaced by a calculator.
Upon acceptance, we will provide a video demo of the interface.

\subsection{Impact of Pairwise Comparisons} \label{appendix:section:comparison}

When running our \interface{} user study, two-thirds of our users are assigned to an experimental group where they complete pairwise comparisons and then execute plans.
The final third are assigned to a group where they do not complete a pairwise comparison, but have the option to switch between plans during plan execution.
Our hope was that this group could form another feedback signal for predicting helpfulness, but we found the feature was often unused---leaving it for future exploration.
We omit all instances where users decided to swap plans during plan execution in our experiments.

However, when users are assigned to the swap group and do not swap between plans, we can measure their accuracy and execution time to study the impact of completing or not completing pairwise comparisons.
We find no clear differences between the distribution of average accuracies and execution times between these users (Figure~\ref{fig:no_swap}), suggesting completing pairwise comparisons has little impact on problem-solving success in \interface{}.

\subsection{ReACT Implementation} \label{appendix:section:react}

 We base our implementation of ReACT on the original framework \cite{yao2023react}, which iteratively runs a three-step protocol of reasoning, acting, and observing.
 The prompts we use for ReACT are in Prompt~\ref{prompt:react_math} for math and Prompt~\ref{prompt:react_trivia} for trivia.
 We use GPT-4o \cite{hurst2024gpt} with $0$ temperature.
 The model was allocated $\sim8$ hours to run on CPU only.
 All results are reported from three runs.

\subsection{Reward Model Implementation} \label{appendix:section:reward_model}

All reward models were implemented according to their official Huggingface code.\footnote{https://huggingface.co/spaces/allenai/reward-bench}
Each model was allocated $\sim3$ hours.
Nemotron was implemented with NVIDIA's official API (CPU-only).\footnote{https://build.nvidia.com/nvidia/llama-3\_1-nemotron-70b-reward}
Other reward models use one NVIDIA:RTXA6000.
All reward model predictions are based on a single run.
Our prompt for the GPT-4o judge is in Prompt~\ref{prompt:gpt_judge} and for all other reward models in Prompt~\ref{prompt:reward_model}.

\subsection{Further Helpfulness Agreement Analysis} \label{appendix:section:agreement}

We now provide further analysis on the agreement of helpfulness signals (\cref{subsection:agreement}).
In Tables~\ref{table:full_agreement_math} and \ref{table:full_agreement_trivia}, we extend our results in Table~\ref{table:agreement} to form full $10 \times 10$ agreement matrices on math and trivia, respectively, to capture \reward{} agreement; \reward{}s have high agreement with each other---often above $80\%$---so they learn similar notions of helpfulness, likely because they have similar training data \cite{lambert2024rewardbench}.

Further, to ensure our \irt{} metric (\cref{subsection:irt}) is not the only reason helpfulness conflicts with preferences, we replicate the agreement analysis in \cref{subsection:agreement} but using accuracy and execution time alone to identify which plan is more helpful.
We also implement a simple average, which first uses average accuracy to denote helpful plans, and then execution time as a tie-break.
Our findings are consistent in math (Figure~\ref{fig:math_help_measures}) and trivia (Figure~\ref{fig:trivia_help_measures}): preferences do not accurately predict which plans actually help users, regardless of whether helpfulness is defined by \irt{}, accuracy, execution time, or averages.

\subsection{\irt{} Analysis} \label{appendix:section:irt} \label{subsection:irt_eval}

Using best practices of evaluating metrics \cite{saxon2024who, shankar2024validates}, we validate our \irt model by studying its convergence, assessing its generalization, interpreting discriminability and skill, ablating our design, and verifying difficulty correlates with accuracy and time as we expect.

\subsubsection{Convergence} \label{appendix:subsection:convergence}

To ensure we have trained \irt{} for sufficient epochs, we first study its convergence.
We find the model quickly converges to modeling the observed data (Figure~\ref{fig:loss}) and reaches low R-hat values and high Effective Sample Sizes (Figure~\ref{fig:converge}), so the model converges across our five chains.
Further, our five chains perfectly agree on which plan in a pair is more helpful (\cref{subsection:agreement}), with Fleiss's $\kappa=1.0$ \cite{fleiss1971measuring}, so \irt{} consistently discerns helpfulness.

\subsubsection{Generalization} \label{appendix:subsection:generalization}

While we primarily use \irt{} to capture helpfulness, we still test its generalization, seeing how much it overfits to our data.
To do this, we train \irt{} on the first $80\%$ of every user's execution history in \interface{}, and check how well it models the observed, held-out accuracy and execution time.
The model has only minor drops in log-likelihood on accuracy ($\sim5\%$), showing it effectively generalizes to user accuracy on new items (Figure~\ref{fig:generalization}).
Log-likelihood does drop more on execution time ($\sim75\%$), but this is to be expected, as predicting response time is generally difficult \cite{ratcliff1978theory}.

\subsubsection{Parameter Interpretations} \label{appendix:subsection:parameters}

While we primarily study difficulty $\beta_i$ in our \irt{} model, the other parameters---item discriminability $\gamma_i$ and player skill $\theta_j$---can also give insights into how users interact with plans in \interface{}.

Discriminability $\gamma_i$ captures how well plans discern between low-skill and high-skill players. In Tables~\ref{table:discrim_math} and \ref{table:discrim_trivia}, we show plans with the highest gap in discriminability for math and trivia, respectively.
In math, plans with higher $\gamma_i$ tend to be longer; it requires more skill to solve problems accurately and quickly with longer plans, likely because excess steps naturally slow players down.
In trivia, plans with higher $\gamma_i$ have unconventional and complex steps: asking users to search for timelines, use self-verification, and follow complex instructions like ``Ascertain'', so only stronger problem-solvers can handle this more difficult level of guidance.

To ensure player skill $\theta_j$ correlates with downstream task success in \interface{}, we plot each $\theta_j$ for player $j$ against their average accuracy, execution time, and number of questions seen.
As expected, players with higher $\theta_j$ tend to be more accurate with lower execution time.
Further, players with higher $\theta_j$ tend to answer more questions, aligning with our results in \cref{subsection:plans_help}, confirming that users become more successful problem-solvers as they keep interacting with plans in \interface{}.

\subsubsection{Ablations} \label{appendix:subsection:ablations}

We ablate two key parts of our model: 1) modeling player skills (Eq.~\ref{equation:prior}); and 2) using base time $\mu_{}$ (Eq.~\ref{equation:base_time}).
Both steps improve the prediction of~our observed data (Figure~\ref{fig:ablation}), with the removal of (1) having the largest drop---showing the need for capturing individual skill to measure helpfulness.

\subsubsection{Helpfulness Interpretation} \label{appendix:subsection:helpfulness_interpretation}

Next, we check \irt{} helpfulness scores (i.e. negative difficulty) behave as intended. 
As true~helpfulness increases, the average accuracy of players on plans tends to rise and average log-time tends to drop (Figure~\ref{fig:trends}), matching our expectation: helpful plans imply accurate, efficient problem-solving \cite{Wang2024PlanningCUA}.
In Appendix~\ref{appendix:section:irt}, we further test our model's convergence and generalization.

\subsection{Qualitative Trace Analysis Details} \label{appendix:section:qual}

To explain how we inferred each trace error type in \cref{subsection:human_agent_trace}, we show examples for each type in math:
\begin{itemize*}
    \item \textbf{Step Error:} A question asks for the average number of branches per foot in several trees.
    The plan tells the user to incorrectly divide the average number of branches by the average height, but this gives the ratio of averages, not the average ratio as the question intended.
    \item \textbf{Ambiguous:} One plan step says ``Subtract the number of spots on Jean's upper torso from the result of Step 1 to find the spots on her sides'' which is meant to convey two computations: ``1) Subtract the spots''; and ``2) find the spots on her sides'', but users interpreted ``to find'' in the step as being the same computation.
    \item \textbf{Execution Error:} One plan asks users to find ``the cost for Jessica's bracelets by multiplying the number of letters in her name by the cost per bracelet'', but the user did $6(2) = 12$, instead of $7(2)=14$, as ``Jessica'' has $7$ letters.
    \item \textbf{Ignored:} One user left all subanswers blank and did not use the calculator (i.e. did calculations in their head), making it impossible to diagnose where they erred in the trace. 
    \item \textbf{Mistake:} GSM8k answers must be rounded to the nearest integer. One user got the answer $81.78$, but instead of rounding to $82$ (the right answer), they submitted $81.78$, leading to an incorrect final response. The user immediately fixed this mistake on their next attempt.
\end{itemize*}

\subsection{User Study Compensation Details}

Part of our compensation for this study was in the form of university extra credit, which if not properly handled, could pose undue pressure on low-performing students.
To address this, we discussed guidelines with the university professors to ensure students were not coerced into participating in our study. Concretely, we provided an alternative assignment of equal difficulty (a coding assignment for a CS class, a reading quiz in a visualization class) to ensure participants could still obtain extra credit if they did not want to participate in our study. Further, we note that the extra credit assignment only required students to answer 25 questions. For most students, this took $\sim$30 minutes; the assignment was also administered over five weeks, including a break period during the Spring term, providing students ample time to complete the study. 

Regarding monetary compensation, we explicitly set the maximum time per question in our study to 180 seconds, so even if a user failed to answer every question in our study, they would still be compensated at a rate of 20 USD/hr---well above our region’s minimum wage. The lowest average execution time of any user in our study was 168 seconds, so this user obtained 21.43 USD/hr.

\subsection{``A Good Man is Hard to Find'' Answers} \label{appendix:story}

If you have already looked for (or found) our references to ``A Good Man is Hard to Find'' by Flannery O'Connor, this section reveals all six of them.

The story concerns a family taking a road trip to Florida (subtly alluded to with the term ``Drive'' in \cref{subsection:plans_help}) and after getting into a car accident, they meet an escaped criminal named The Misfit (thus, ``Escape'' in \cref{subsection:agreement} and ``Misfit Plans'' in \cref{section:qualitative}).
This dialogue has many notable quotes, like ``\textit{She would of been a good woman \dots if it had been somebody there to
shoot her every minute of her life}'' (mirrored in \cref{subsection:qual_diff}) and the last line of ``\textit{It's no real pleasure in life.}'' (mirrored in \cref{subsection:human_agent_trace}).
Lastly, as a silly piece of trivia\footnote{https://www.naqt.com/}, O'Connor details a watermelon with the initials ``E.A.T.'' carved into it (reflected in \cref{subsection:regression}).
We would be impressed if you got them all!

\input{figures/planorama_math}
\input{appendix/dataset}
\input{appendix/no_swap}
\input{appendix/loss}
\input{appendix/converge}
\input{appendix/generalization}
\input{appendix/help_measures}
\input{appendix/skill}
\input{data/ablation}
\input{data/trend}
\input{appendix/full_agreement}
\input{appendix/discrim_math}
\input{appendix/discrim_trivia}

\clearpage
\clearpage
\input{appendix/prompts}

%% file: figures/planorama_math.tex
\begin{figure*}
    \centering
    \fbox{\includegraphics[width=\linewidth]{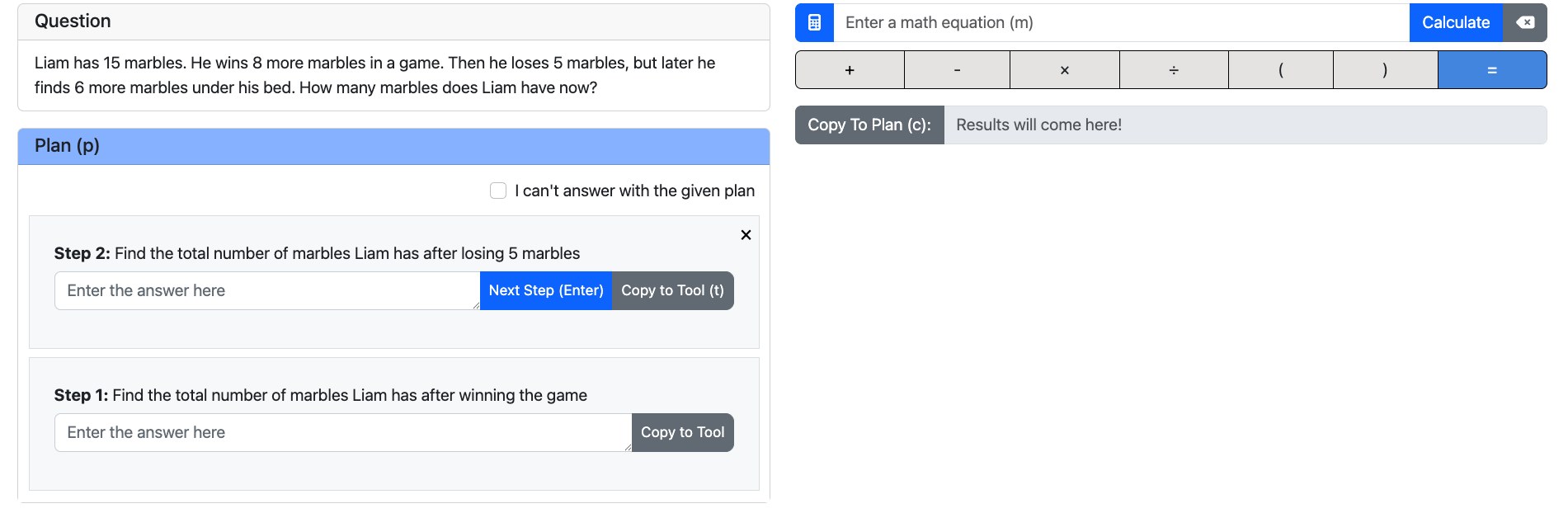}}
    \caption{\small Overview of the \interface{} interface for answering math questions. The interface mirrors Figure~\ref{fig:planorama}, but users have access to a calculator (right) rather than search.}
    \label{fig:planorama_math}
\end{figure*}

%% file: appendix/dataset.tex
\begin{table*}
\small
\centering
\setlength{\tabcolsep}{3.7pt}
\begin{tabular}{@{}ccccccc@{}}
\toprule
       & \# $q$ & Source(s)      & Avg Words Per $q$ & \# Steps / $p$ & Avg Executions / $p$ & Avg Comparisons / $p$ \\ \midrule
Math   & 150          & GSM8k (150)          & 50.48                  & 3.11            & 8.99                 & 9.91                  \\
Trivia & 150          & MuSiQue (10), MQuAKE (140) & 18.39                  & 2.85            & 7.86                 & 9.91                  \\ \bottomrule
\end{tabular}
\caption{\small Summary of the \interface{} dataset. We use 150 math and trivia questions mainly from GSM8k and MQuAKE. Our questions are supported by multi-step plans, typically 2-3 steps with high-level guidance for answering the question. We collect rich feedback from our users, with an average of 8.99 and 7.86  execution traces per plan, and 9.91 pairwise comparisons per plan.}
\label{appendix:table:dataset}
\end{table*}

%% file: appendix/no_swap.tex
\begin{figure*}
    \centering
    \fbox{\includegraphics[width=\linewidth]{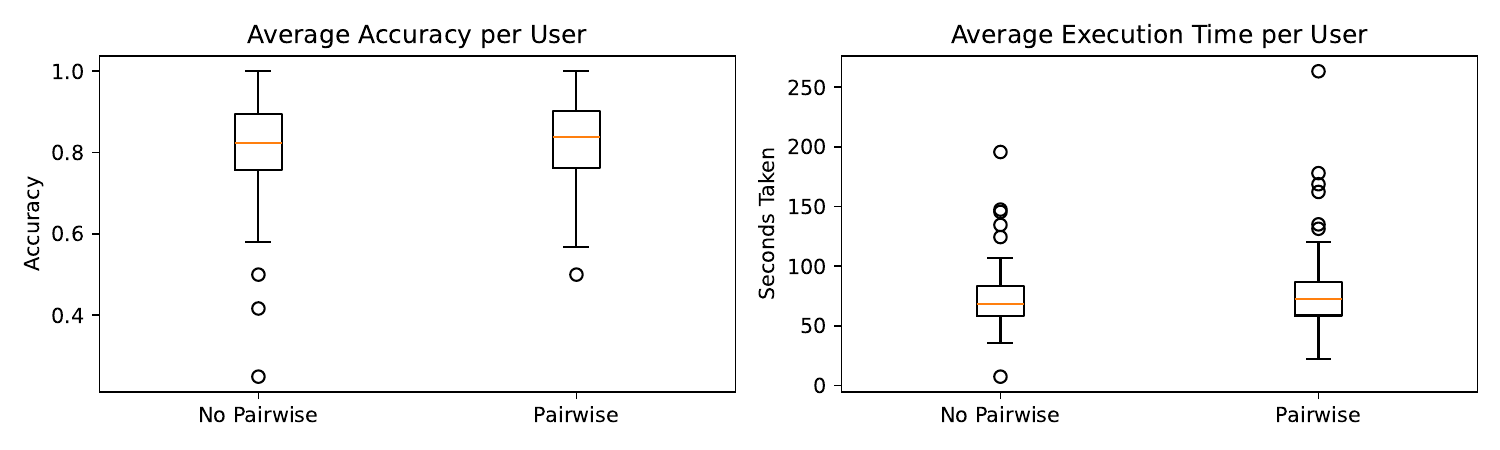}}
    \caption{\small We find little differences in the distribution of average accuracies and execution times between users who complete and do not complete pairwise comparisons, suggesting it does not have much impact on problem-solving success.}
    \label{fig:no_swap}
\end{figure*}

%% file: appendix/loss.tex
\begin{figure*}
    \centering
    \fbox{\includegraphics[width=\linewidth]{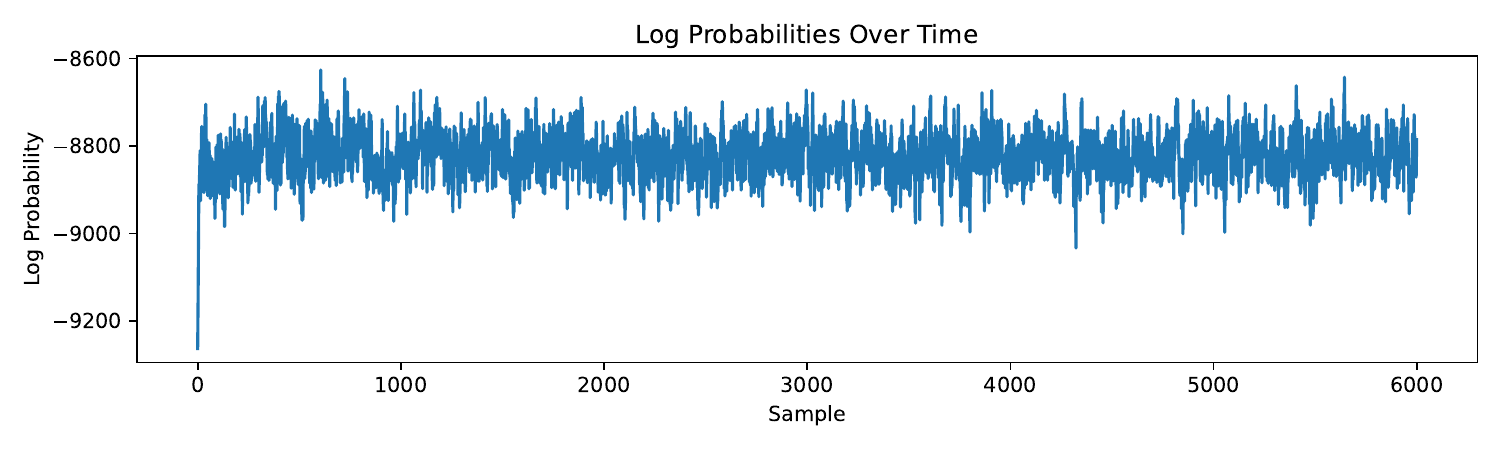}}
    \caption{\small Plot of our \irt{} model's log-probability of predicting observed data. Our model quickly converges after a few samples, showing it adequately fits to our observed accuracy and execution time.}
    \label{fig:loss}
\end{figure*}

%% file: appendix/converge.tex
\begin{figure*}
    \centering
    \fbox{\includegraphics[width=\linewidth]{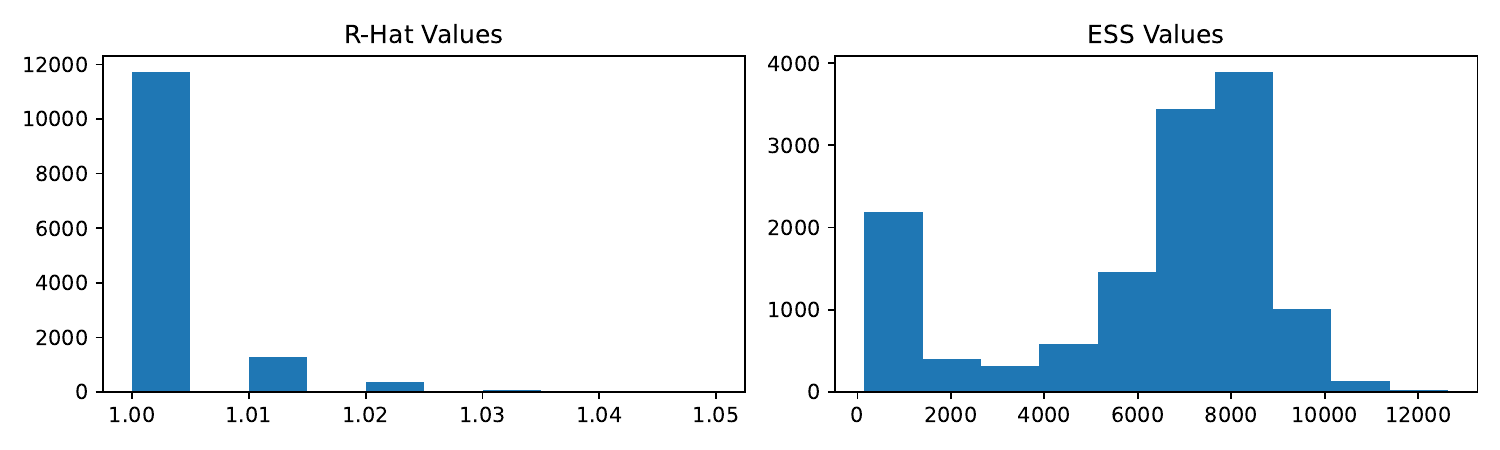}}
    \caption{\small Distribution of R-hat and Effective Sample Size (ESS) values for our \irt{} model. R-Hat is always under 1.05 and the majority of ESS's are in the thousands, indicating strong convergence across our five chains.}
    \label{fig:converge}
\end{figure*}

%% file: appendix/generalization.tex
\begin{figure*}
    \centering
    \fbox{\includegraphics[width=\linewidth]{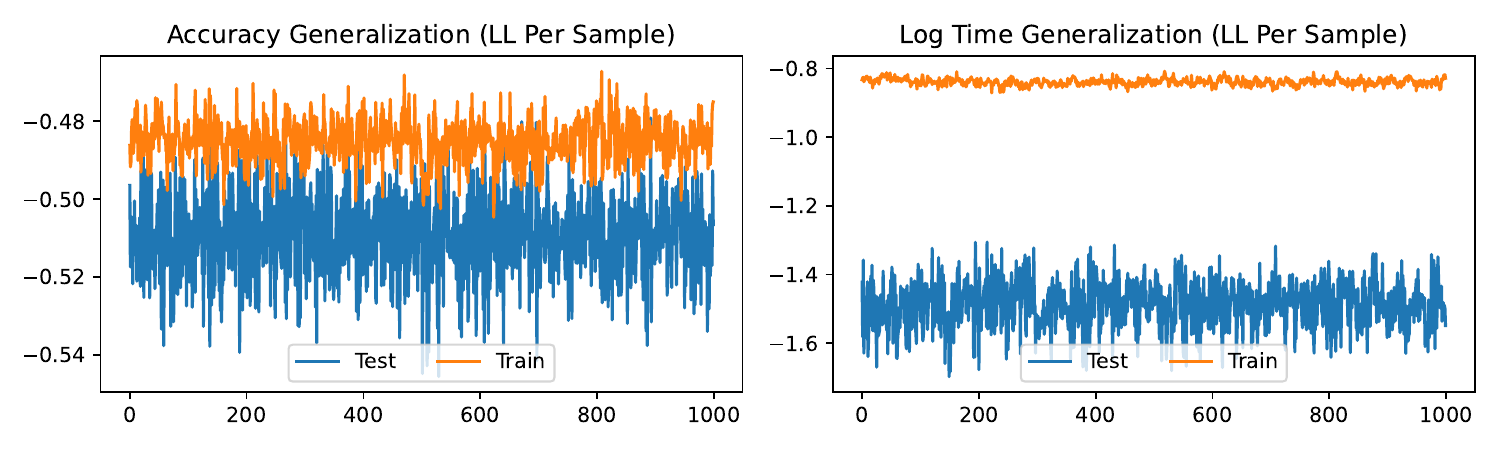}}
    \caption{\small Generalization of \irt{} when predicting accuracy and execution time after being trained on the first 80\% of the user's interactions in \interface{}. The model effectively generalizes to predict accuracy, but struggles more with execution time.}
    \label{fig:generalization}
\end{figure*}

%% file: appendix/help_measures.tex
\begin{figure*}
    \centering
    \fbox{\includegraphics[width=\linewidth]{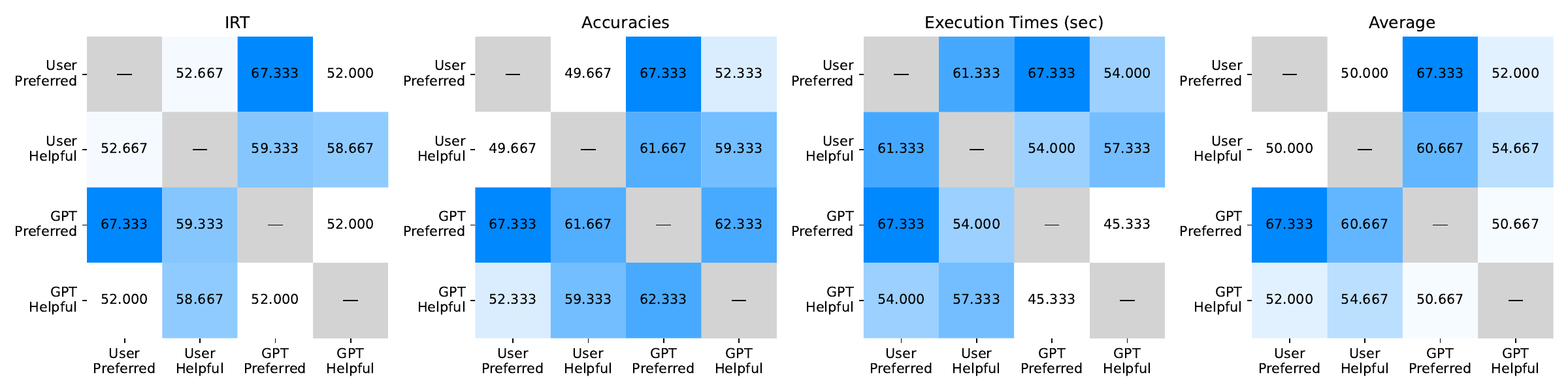}}
    \caption{\small User/model perceived and true helpfulness agreement on \textbf{math} (Table~\ref{table:agreement}, left) based on whether \irt{}, accuracy,  time, or average accuracy/time dictates which plan is helpful. In every case, nothing accurately predicts what is truly helpful for users.}
    \label{fig:math_help_measures}
\end{figure*}

\begin{figure*}
    \centering
    \fbox{\includegraphics[width=\linewidth]{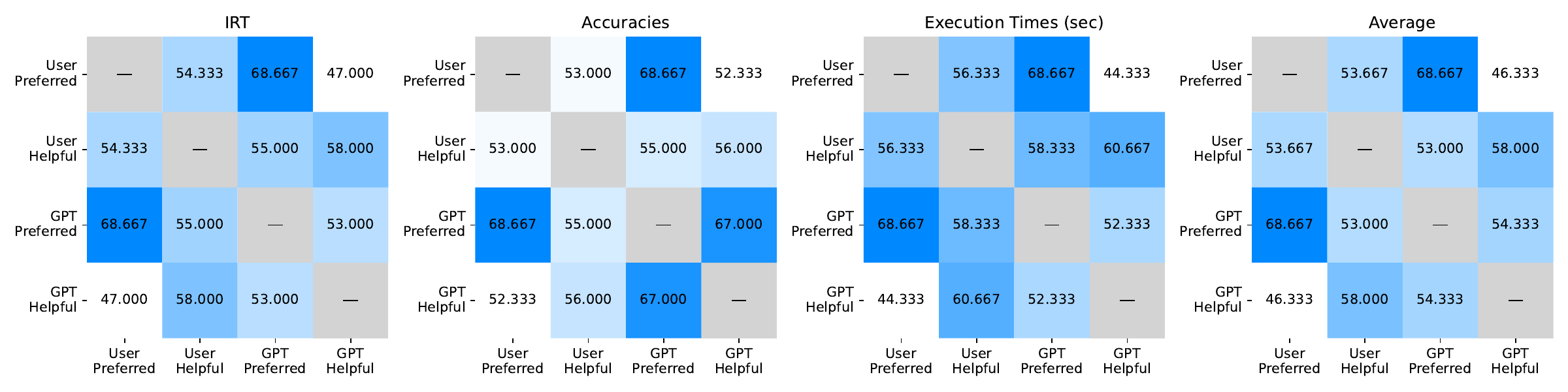}}
    \caption{\small User/model perceived and true helpfulness agreement on \textbf{trivia} (Table~\ref{table:agreement}, right) based on whether \irt{}, accuracy, time, or average accuracy/time dictates which plan is  helpful. In every case, nothing accurately predicts what is truly helpful for users.}
    \label{fig:trivia_help_measures}
\end{figure*}

%% file: appendix/skill.tex
\begin{figure*}
    \centering
    \fbox{\includegraphics[width=\linewidth]{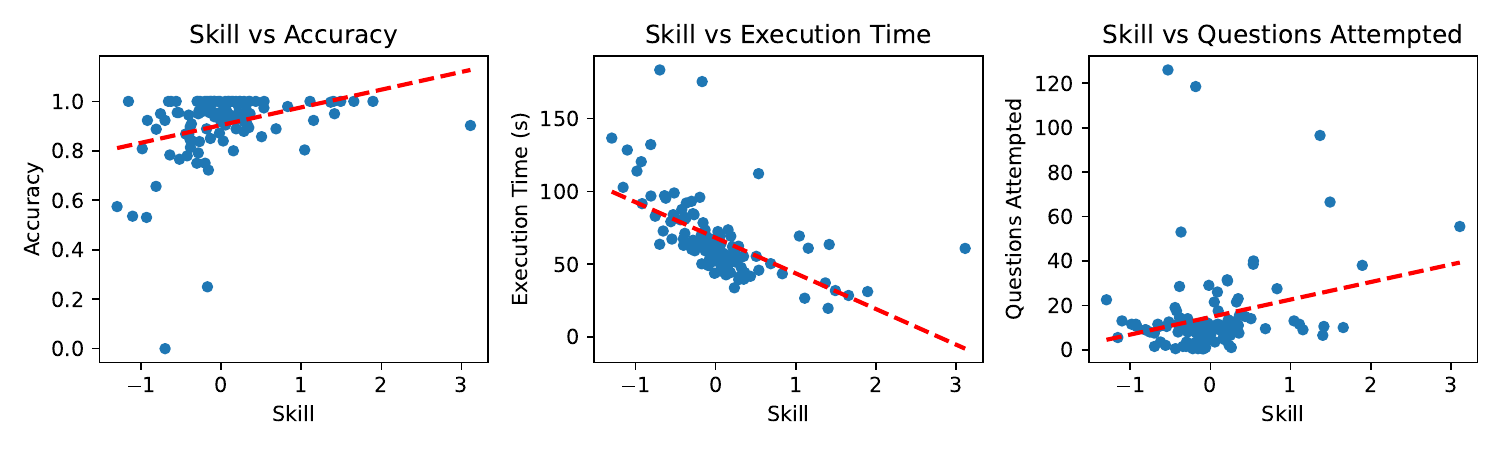}}
    \caption{\small Correlation between player skill and \interface{} interactions. Players with higher skill are typically more accurate, need less excecution time, and attempt more questions, aligning with our intuition.}
    \label{fig:skill}
\end{figure*}

%% file: data/ablation.tex
\begin{figure*}
    \centering
    \fbox{\includegraphics[width=\linewidth]{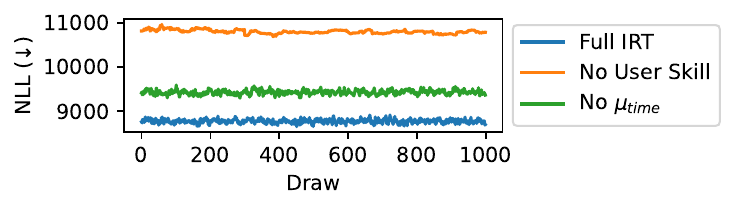}}
    \caption{\small Ablating player skill (Eq.~\ref{equation:prior}) and mean time priors (Eq.~\ref{equation:base_time}) degrade \irt{}'s observed data predictions (negative log-likelihood), showing both better model true plan helpfulness.}
    \label{fig:ablation}
\end{figure*}

%% file: data/trend.tex
\begin{figure*}
    \centering
\fbox{\includegraphics[width=\linewidth]{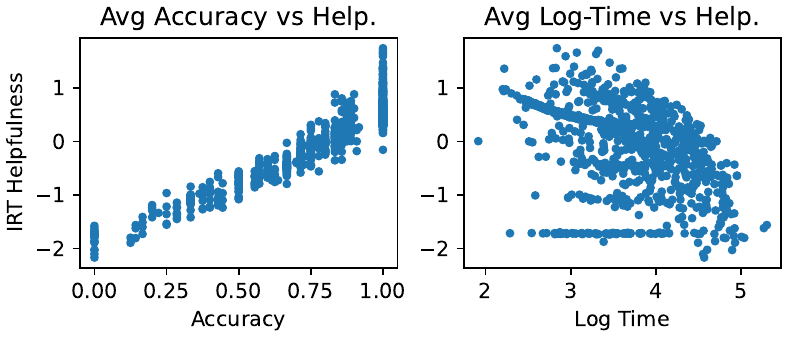}}
    \caption{\small As \irt{}'s true helpfulness metric rises, per-player accuracy rises and log-time drops, matching our expectations.}
    \label{fig:trends}
\end{figure*}

%% file: appendix/full_agreement.tex
\begin{table*}
\small
\centering
\setlength{\tabcolsep}{3pt}
\begin{tabular}{@{}c|cccc|cccccc@{}}
\toprule
\textbf{Proxy} & User Judge & User Help & GPT Judge & GPT Help & QRM    & GRM    & Skywork & Nemotron & InternLM2 & ArmoRM \\ \midrule
User Judge     & ---        & ---       & ---       & ---      & ---    & ---    & ---     & ---      & ---       & ---    \\
User Help      & \cellcolor[RGB]{189,224,255} 52.000 \strut     & ---       & ---       & ---      & ---    & ---    & ---     & ---      & ---       & ---    \\ \midrule
GPT Judge      & \cellcolor[RGB]{113,188,255} 67.333 \strut     & \cellcolor[RGB]{156,208,255} 58.667 \strut    & ---       & ---      & ---    & ---    & ---     & ---      & ---       & ---    \\
GPT Help       & \cellcolor[RGB]{172,216,255} 55.333 \strut     & \cellcolor[RGB]{162,211,255} 57.333 \strut    & \cellcolor[RGB]{186,222,255} 52.667 \strut    & ---      & ---    & ---    & ---     & ---      & ---       & ---    \\ \midrule
QRM            & \cellcolor[RGB]{149,205,255} 60.000 \strut     & \cellcolor[RGB]{169,214,255} 56.000 \strut     & \cellcolor[RGB]{89,177,255} 72.000 \strut    & \cellcolor[RGB]{236,246,255} 42.667 \strut   & ---    & ---    & ---     & ---      & ---       & ---    \\
GRM            & \cellcolor[RGB]{182,220,255} 53.333 \strut     & \cellcolor[RGB]{176,218,255} 54.667 \strut    & \cellcolor[RGB]{119,191,255} 66.000 \strut    & \cellcolor[RGB]{255,255,255} 38.667 \strut   & \cellcolor[RGB]{43,156,255} 81.333 \strut & ---    & ---     & ---      & ---       & ---    \\
Skywork        & \cellcolor[RGB]{116,190,255} 66.667 \strut     & \cellcolor[RGB]{192,225,255} 51.333 \strut    & \cellcolor[RGB]{93,179,255} 71.333 \strut    & \cellcolor[RGB]{232,244,255} 43.333 \strut   & \cellcolor[RGB]{0,136,255} 90.000 \strut & \cellcolor[RGB]{46,157,255} 80.667 \strut & ---     & ---      & ---       & ---    \\
Nemotron       & \cellcolor[RGB]{179,219,255} 54.000 \strut     & \cellcolor[RGB]{149,205,255} 60.000 \strut    & \cellcolor[RGB]{116,190,255} 66.667 \strut    & \cellcolor[RGB]{249,252,255} 40.000 \strut   & \cellcolor[RGB]{48,158,255} 80.000 \strut & \cellcolor[RGB]{82,174,255} 73.333 \strut & \cellcolor[RGB]{79,172,255} 74.000 \strut  & ---      & ---       & ---    \\
InternLM2      & \cellcolor[RGB]{162,211,255} 57.333 \strut     & \cellcolor[RGB]{162,211,255} 57.333 \strut    & \cellcolor[RGB]{96,180,255} 70.667 \strut    & \cellcolor[RGB]{242,248,255} 41.333 \strut   & \cellcolor[RGB]{36,152,255} 82.667 \strut & \cellcolor[RGB]{76,171,255} 74.667 \strut & \cellcolor[RGB]{65,166,255} 76.667 \strut  & \cellcolor[RGB]{48,158,255} 80.000 \strut   & ---       & ---    \\
ArmoRM         & \cellcolor[RGB]{166,213,255} 56.667 \strut     & \cellcolor[RGB]{166,213,255} 56.667 \strut    & \cellcolor[RGB]{109,186,255} 68.000 \strut    & \cellcolor[RGB]{252,253,255} 39.333 \strut   & \cellcolor[RGB]{59,163,255} 78.000 \strut & \cellcolor[RGB]{32,151,255} 83.333 \strut & \cellcolor[RGB]{63,165,255} 77.333 \strut  & \cellcolor[RGB]{46,157,255} 80.667 \strut   & \cellcolor[RGB]{53,160,255} 79.333 \strut    & ---    \\ \bottomrule
\end{tabular}
\caption{\small Full agreement analysis from Table~\ref{table:agreement} on math questions. As expected, \reward{}s have high agreement with each other, showing that they all learn similar notions of helpfulness.} \label{table:full_agreement_math}
\end{table*}

\begin{table*}
\small
\centering
\setlength{\tabcolsep}{3pt}
\begin{tabular}{@{}c|cccc|cccccc@{}}
\toprule
\textbf{Proxy} & User Judge & User Help & GPT Judge & GPT Help & QRM    & GRM    & Skywork & Nemotron & InternLM2 & ArmoRM \\ \midrule
User Judge     & ---        & ---       & ---       & ---      & ---    & ---    & ---     & ---      & ---       & ---    \\
User Help      & \cellcolor[RGB]{161,211,255} 55.667 \strut     & ---       & ---       & ---      & ---    & ---    & ---     & ---      & ---       & ---    \\ \midrule
GPT Judge      & \cellcolor[RGB]{102,183,255} 68.667 \strut     & \cellcolor[RGB]{158,209,255} 56.333 \strut    & ---       & ---      & ---    & ---    & ---     & ---      & ---       & ---    \\
GPT Help       & \cellcolor[RGB]{191,225,255} 49.000 \strut     & \cellcolor[RGB]{129,196,255} 62.667 \strut    & \cellcolor[RGB]{167,213,255} 54.333 \strut    & ---      & ---    & ---    & ---     & ---      & ---       & ---    \\ \midrule
QRM            & \cellcolor[RGB]{114,189,255} 65.667 \strut     & \cellcolor[RGB]{179,220,255} 51.333 \strut    & \cellcolor[RGB]{158,209,255} 56.333 \strut    & \cellcolor[RGB]{246,250,255} 36.667 \strut   & ---    & ---    & ---     & ---      & ---       & ---    \\
GRM            & \cellcolor[RGB]{121,192,255} 64.333 \strut     & \cellcolor[RGB]{179,220,255} 51.333 \strut    & \cellcolor[RGB]{155,208,255} 57.000 \strut    & \cellcolor[RGB]{228,242,255} 40.667 \strut   & \cellcolor[RGB]{32,151,255} 84.000 \strut & ---    & ---     & ---      & ---       & ---    \\
Skywork        & \cellcolor[RGB]{112,188,255} 66.333 \strut     & \cellcolor[RGB]{171,215,255} 53.333 \strut    & \cellcolor[RGB]{146,204,255} 59.000 \strut    & \cellcolor[RGB]{231,243,255} 40.000 \strut   & \cellcolor[RGB]{0,136,255} 91.333 \strut & \cellcolor[RGB]{36,152,255} 83.333 \strut & ---     & ---      & ---       & ---    \\
Nemotron       & \cellcolor[RGB]{143,202,255} 59.667 \strut     & \cellcolor[RGB]{183,221,255} 50.667 \strut    & \cellcolor[RGB]{170,215,255} 53.667 \strut    & \cellcolor[RGB]{255,255,255} 34.667 \strut   & \cellcolor[RGB]{30,150,255} 84.667 \strut & \cellcolor[RGB]{24,147,255} 86.000 \strut & \cellcolor[RGB]{32,151,255} 84.000 \strut  & ---      & ---       & ---    \\
InternLM2      & \cellcolor[RGB]{137,199,255} 61.000 \strut     & \cellcolor[RGB]{174,217,255} 52.667 \strut    & \cellcolor[RGB]{179,219,255} 51.667 \strut    & \cellcolor[RGB]{240,248,255} 38.000 \strut   & \cellcolor[RGB]{32,151,255} 84.000 \strut & \cellcolor[RGB]{27,148,255} 85.333 \strut & \cellcolor[RGB]{36,152,255} 83.333 \strut  & \cellcolor[RGB]{6,138,255} 90.000 \strut   & ---       & ---    \\
ArmoRM         & \cellcolor[RGB]{143,202,255} 59.667 \strut     & \cellcolor[RGB]{177,218,255} 52.000 \strut    & \cellcolor[RGB]{173,216,255} 53.000 \strut    & \cellcolor[RGB]{219,238,255} 42.667 \strut   & \cellcolor[RGB]{65,166,255} 76.667 \strut & \cellcolor[RGB]{30,150,255} 84.667 \strut & \cellcolor[RGB]{63,165,255} 77.333 \strut  & \cellcolor[RGB]{56,162,255} 78.667 \strut   & \cellcolor[RGB]{41,155,255} 82.000 \strut    & ---    \\ \bottomrule
\end{tabular}
\caption{\small Full agreement analysis from Table~\ref{table:agreement} on trivia questions. As expected, \reward{}s have high agreement with each other, showing that they all learn similar notions of helpfulness.} \label{table:full_agreement_trivia}
\end{table*}

%% file: appendix/discrim_math.tex
\begin{table*}
\small
\centering
\begin{tabular}{@{}p{5cm} p{5cm} p{5cm}@{}}
\toprule
\textbf{Question} & \textbf{High Discriminability} & \textbf{Low Discriminability} \\ 
\midrule
Carly is a pet groomer. Today, her task was trimming the four nails on each of the dogs’ paws. She trimmed 164 nails, but three of the dogs had only three legs. How many dogs did Carly work on? 
& 
\begin{minipage}[t]{\linewidth}
\begin{enumerate*}
    \item Calculate how many nails are trimmed for a dog with three legs.
    \item Determine the total number of nails missing from all three-legged dogs.
    \item Find out the total number of nails as if all dogs had four legs.
    \item Calculate the total number of dogs by dividing total four-legged nails by nails per four-legged dog.
\end{enumerate*}
\end{minipage}
& 
\begin{minipage}[t]{\linewidth}
\begin{enumerate*}
    \item Estimate the number of four-legged dogs using the total number of nails.
    \item Subtract the number of three-legged dogs to find four-legged dogs.
    \item Add the three-legged dogs back to get the total number of dogs.
\end{enumerate*}
\end{minipage}
\\
\midrule
Two sisters, Elizabeth and Margareth, bought beads. Elizabeth bought 1 pack of red and 2 packs of clear beads, while Margareth bought 3 packs of blue and 4 packs of red beads. How many more beads does one sister have than the other, if each pack contains 20 beads?
& 
\begin{minipage}[t]{\linewidth}
\begin{enumerate*}
    \item Start by calculating the total number of beads in one pack of red beads.
    \item Use that to calculate the total for Elizabeth's red and clear beads.
    \item Calculate the total number of beads in Margareth's blue packs.
    \item Calculate the total number of beads in Margareth's red packs.
    \item Calculate the total beads for Margareth and subtract Elizabeth's total beads to find the difference.
\end{enumerate*}
\end{minipage}
& 
\begin{minipage}[t]{\linewidth}
\begin{enumerate*}
    \item Figure out how many beads Elizabeth bought by multiplying number of packs by beads per pack.
    \item Do the same for Margareth.
    \item Find the difference by subtracting the smaller total from the larger one.
\end{enumerate*}
\end{minipage}
\\
\midrule
Parker wants to find out what the average percentage of kernels that pop in a bag is. In the first bag he makes, 60 kernels pop and the bag has 75 kernels. In the second bag, 42 kernels pop and there are 50 in the bag. In the final bag, 82 kernels pop and the bag has 100 kernels. What is the average percentage?
& 
\begin{minipage}[t]{\linewidth}
\begin{enumerate*}
    \item Find the percentage popped for the first bag.
    \item Repeat for the second and third bags.
    \item Add the three percentages.
    \item Divide by 3 to find the average.
\end{enumerate*}
\end{minipage}
& 
\begin{minipage}[t]{\linewidth}
\begin{enumerate*}
    \item Find the total number of popped kernels across all bags.
    \item Find the total number of kernels across all bags.
    \item Divide the popped kernels by the total kernels and multiply by 100.
\end{enumerate*}
\end{minipage}
\\
\bottomrule
\end{tabular}
\caption{\small Comparison of the three plan pairs in math questions with the highest gap in discriminability.}
\label{table:discrim_math}
\end{table*}

%% file: appendix/discrim_trivia.tex
\begin{table*}
\small
\centering
\begin{tabular}{@{}p{5cm}p{5cm}p{5cm}@{}}
\toprule
\textbf{\centering\arraybackslash Question} & \textbf{\centering\arraybackslash High Discriminability} & \textbf{\centering\arraybackslash Low Discriminability} \\ 
\midrule
How many stars are on the flag of the country where the spouse of the performer of Wrecking Ball is a citizen of?
& 
\begin{minipage}[t]{\linewidth}
\begin{enumerate*}
    \item Identify a song by a known artist that matches Wrecking Ball's release timeline.
    \item Determine the country of citizenship for the known spouse of the artist from Step 1.
    \item Find the number of stars on the flag of the country from Step 2.
\end{enumerate*}
\end{minipage}
& 
\begin{minipage}[t]{\linewidth}
\begin{enumerate*}
    \item Find the performer of the song Wrecking Ball.
    \item Identify the spouse of the performer from Step 1.
    \item Determine the country of citizenship for the spouse from Step 2.
    \item Find the number of stars on the flag of the country from Step 3.
\end{enumerate*}
\end{minipage}
\\
\midrule
Who built the castle named after the city with an institution that educated the author of Species Plantarum?
& 
\begin{minipage}[t]{\linewidth}
\begin{enumerate*}
    \item Find the name of the author of Species Plantarum.
    \item Determine the institution that educated the author identified in Step 1.
    \item Identify the city where the institution found in Step 2 is located.
    \item Ascertain the builder of the castle that carries the name of the city found in Step 3.
\end{enumerate*}
\end{minipage}
& 
\begin{minipage}[t]{\linewidth}
\begin{enumerate*}
    \item Identify the city that is home to an institution where the author of Species Plantarum was educated.
    \item Find the name of the castle that shares its name with the city identified in Step 1.
    \item Determine the builder of the castle identified in Step 2.
\end{enumerate*}
\end{minipage}
\\
\midrule
What is the city did the spouse of Nicolae Ceaușescu move to after receiving elementary education?
& 
\begin{minipage}[t]{\linewidth}
\begin{enumerate*}
    \item Identify Nicolae Ceaușescu's spouse.
    \item Find the city the spouse moved to after elementary education.
    \item Check if the city in Step 2 is the same as the city where the spouse received elementary education.
    \item If the cities are different, find the city the spouse moved to after receiving elementary education.
\end{enumerate*}
\end{minipage}
& 
\begin{minipage}[t]{\linewidth}
\begin{enumerate*}
    \item Find the city where Nicolae Ceaușescu's spouse went to elementary school.
    \item Determine the city the spouse moved to after Step 1.
\end{enumerate*}
\end{minipage}
\\
\bottomrule
\end{tabular}
\caption{\small Comparison of the three plan pairs in trivia questions with the highest gap in discriminability.}
\label{table:discrim_trivia}
\end{table*}

%% file: appendix/prompts.tex
\hypersetup{
    linkcolor=white, 
    citecolor=white, 
    urlcolor=white 
}

\begin{prompt}[title={Prompt \thetcbcounter: Math Plan Generation Prompt (\cref{subsection:dataset})}, label=prompt:math_plan]
I will give you a math question, and your goal is to come up with two diverse plans with instructions that can help someone answer the question. The plans should lead to the correct answer, but more importantly should be highly optimized for speed. Each step of the plan should not reveal the answer, intermediate computations, any specific numbers in the input question, or how to exactly perform any calculation. Plans should focus on basic computations rather than setting up complex equations. Each plan should be between 2 and 10 steps. Steps should be high-level, brief, clear, and not include details about how to compute any intermediate values.\\\\
Please format your output as a JSON dictionary with key "plan1” for the first plan and “plan2” for the second plan. The value for each key should be a list of strings with steps on how to answer the question. Each step should be prefixed by its number (e.g. "Step 1:"). Steps should reference outputs form preceding steps using the number of the step.\\\\
Remember, plans should be accurate but highly optimized for speed which should be achieved by minimizing the number of steps, keeping descriptions brief, and using shortcuts that can skip the traditional route of answering the question. The two plans can differ in their overall strategy, specificity, number of steps, and what intermediate information to compute. \\\\
Produce plans for the question: $q$
\end{prompt}

\begin{prompt}[title={Prompt \thetcbcounter: Trivia Plan Generation Prompt (\cref{subsection:dataset})}, label=prompt:trivia_plan]
I will give you a trivia question, and your goal is to come up with two diverse plans with instructions that can help someone answer the question. Do not reference any tools or sources that should be used. The plans should lead to the correct answer, but more importantly should be highly optimized for speed. Each step of the plan should instruct the user to find an intermediate answer. Plans should not reveal the answer or intermediate answers, and can only contain information in the input question. Each plan should be between 2 and 10 steps. Steps should be high-level, brief, self-contained, and cannot include extra information, entities, and knowledge that is not present in the input question.\\ \\
Please format your output as a JSON dictionary with key "plan1” for the first plan and “plan2” for the second plan. The value for each key should be a list of strings with steps on how to answer the question. Each step should be prefixed by its number (e.g. "Step 1:"). Steps should reference outputs form preceding steps using the number of the step.\\ \\
Remember, plans should be accurate but highly optimized for speed which should be achieved by minimizing the number of steps, keeping descriptions brief, and using shortcuts that can skip the traditional route of answering the question. The two plans can differ in their overall strategy, specificity, number of steps, and what intermediate information to search for.\\ \\
Produce plans for the question: $q$
\end{prompt}

\begin{prompt}[title={Prompt \thetcbcounter: GPT-4o Direct Answer Prompt (\cref{subsection:plans_help})}, label=prompt:gpt_answer]
Answer the following question. Give just the answer and no explanation. Format your final answer as "Answer: [insert generated answer]"\\\\Question: $q$
\end{prompt}

\begin{prompt}[title={Prompt \thetcbcounter: GPT-4o Judge Prompt (\cref{subsection:reward_model})}, label=prompt:gpt_judge]
You will be given a question and two step-by-step plans that could help a human user answer the question (Plan A and Plan B). Your goal is to determine which plan would help a human user answer the question more accurately and quickly. Respond with just the letter of the plan.\\
Question: $q$\\
Plan A: $p_A$\\
Plan B: $p_B$\\
More Helpful Plan:

\end{prompt}

\begin{prompt}[title={Prompt \thetcbcounter: Reward Model Prompt (\cref{subsection:reward_model})}, label=prompt:reward_model]
<user>Generate a plan to help me answer this question accurately and quickly: $q$<user>\\
<assistant>$p$<assistant>
\end{prompt}

\clearpage

\begin{prompt}[title={Prompt \thetcbcounter: ReACT Math Prompt (\cref{subsection:agent})}, label=prompt:react_math]
You will be a given a question and series of previous steps, actions, and thoughts, and your task is to generate the next Thought or Action that will lead you to the correct answer of the current step. Thoughts contain reasoning chains that help you decide which action you should call, while Actions are tool calls that can provide external information. The tools you have access to are:\\
- CALCULATE: Given an input equation, returns the result when evaluating the expression\\
- SUBMIT\_STEP: Submit an answer to the step\\\\
All outputs from tool calls will be provided as Observations. You must call SUBMIT\_STEP to answer each step, not to answer the final question. Below is an example of a full reasoning trace:\\
---\\
Question: Liam has 15 marbles. He wins 8 more marbles in a game. Then he loses 5 marbles, but later he finds 6 more marbles under his bed. How many marbles does Liam have now?\\
---\\
Step 1: Find the total number of marbles Liam has after winning the game\\
Thought: To find the total number of marbles Liam has after winning the game, we must add his initial 15 marbles with the 8 marbles he won after the game\\
Action: CALCULATE(15 + 8)\\
Observation: 23\\\\
Thought: We now have the number of marbles Liam has after winning the game, so we can submit 23 as the answer to this step\\
Action: SUBMIT\_STEP(23)\\
Answer to Step 1: 23\\
---\\
Step 2: Find the total number of marbles Liam has after losing 5 marbles\\
Thought: To get the number of marbles Liam has after losing 5 marbles, we must subtract 5 from the 23 marbles in Step 1\\
Action: CALCULATE(23 - 5)\\
Observation: 18 \\\\
Thought: We now have the number of marbles Liam has after he loses 5 of them, so we can submit 18 as the answer to this step\\
Action: SUBMIT\_STEP(18)\\
Answer to Step 2: 18\\
---\\
Step 3: Find the final number of marbles Liam ends up with after finding more marbles under his bed\\
Thought: The final number of marbles Liam has is the 18 marbles from Step 2 plus the 6 more marbles Liam finds under his bed\\
Action: CALCULATE(18 + 6)\\
Observation: 24\\ \\
Thought: We now have the final number of marbles Liam ends up with, so we can submit 24 as the answer to this step\\
Action: SUBMIT\_STEP(24)\\
Answer to Step 3: 24\\ \\
Now, generate the next Thought or Action for the following question: $q$\\
---
\end{prompt}

\clearpage

\begin{prompt}[title={Prompt \thetcbcounter: ReACT Trivia Prompt (\cref{subsection:agent})}, label=prompt:react_trivia]
You will be a given a question and series of previous steps, actions, and thoughts, and your task is to generate the next Thought or Action that will lead you to the correct answer of the current step. Thoughts contain reasoning chains that help you decide which action you should call, while Actions are tool calls that can provide external information. The tools you have access to are:\\
- SEARCH: Given an input search query, returns the title, first paragraph, and most similar context within a relevant Wikipedia page\\
- SUBMIT\_STEP: Submit an answer to the step\\ \\
All outputs from tool calls will be provided as Observations. You must call SUBMIT\_STEP to answer each step, not to answer the final question. Below is an example of a full reasoning trace:\\
---\\
Question: What is the capital of the state that contains the tallest mountain in the United States?\\
---\\
Step 1: Find the tallest mountain in the United States\\
Thought: To find the tallest mountain in the United States, we can search for the right Wikipedia page with this query\\
Action: SEARCH(tallest mountain United States)\\
Observation: <Title>List of mountain peaks of the United States</Title><First Paragraph>This article comprises three sortable tables of major mountain peaks of the United States of America. </First Paragraph><Selected Content>The summit of a mountain or hill may be measured in three main ways:  Download coordinates as:  Of the 100 highest major summits of the United States, only Denali exceeds 6000 meters (19,685 feet) elevation...</Selected Content>\\
Action: SUBMIT\_STEP(Mount Denali)\\
Answer to Step 1: Mount Denali\\
---\\
Step 2: Find the state that contains this mountain\\\\
Thought: To find the state with this mountain, we must find where Mount Denali is located\\
Action: SEARCH(Mount Denali location)\\
Observation: <Title>Denali</Title><First Paragraph>Denali (/de'na:li/), federally designated as Mount McKinley, is the highest mountain peak in North America, with a summit elevation of 20,310 feet (6,190m) above sea level.  It is the tallest mountain in the world from base to peak on land, measuring 18,000ft (5,500m).  With a topographic prominence of 20,194 feet (6,155m) and a topographic isolation of 4,621.1 miles (7,436.9km), Denali is the third most prominent and third-most isolated peak on Earth, after Mount Everest and Aconcagua.  Located in the Alaska Range in the interior of the U.S. state of Alaska, Denali is the centerpiece of Denali National Park and Preserve. </First Paragraph><Selected Content>Denali (/de'na:li/), federally designated as Mount McKinley, is the highest mountain peak in North America, with a summit elevation of 20,310 feet (6,190m) above sea level...</Selected Content>\\\\
Thought: The last observation states that Denali is located in the U.S. state of Alaska, so we can submit Alaska as the answer to this step\\
Action: SUBMIT\_STEP(Alaska)\\
Answer to Step 2: Alaska\\
---\\
Step 3: Find the capital of this state\\\\
Thought: To find the capital of this state, we must find the capital of Alaska, which can be found on its Wikipedia page
Action: SEARCH(Alaska capital)\\
Observation:<Title>Juneau, Alaska</Title><First Paragraph>Juneau ('dZu:noU i JOO-noh; Tlingit: Dzanti K'iheeni ['tsnthI k'I'hi:nI] transl.  Base of the Flounder's River), officially the City and Borough of Juneau, is the capital of the U.S. state of Alaska, located along the Gastineau Channel and the Alaskan panhandle.  Juneau was named the capital of Alaska in 1906, when the government of what was then the District of Alaska was moved from Sitka as dictated by the U.S. Congress in 1900.  On July 1, 1970, the City of Juneau merged with the City of Douglas and the surrounding Greater Juneau Borough to form the current consolidated city-borough, which ranks as the second-largest municipality in the United States by area and is larger than either Rhode Island or Delaware. </First Paragraph><Selected Content>Atop the mountains is the Juneau Icefield, a large ice mass from which about 30 glaciers flow; two of them, the Mendenhall Glacier and the Lemon Creek Glacier, are visible from the local road system... </Selected Content>\\\\
Thought: The title of this page is Juneau, Alaska, so we can submit Juneau as the answer to this step\\
Action: SUBMIT\_STEP(Juneau)\\
Answer to Step 3: Juneau\\
---\\
Now, generate the next Thought or Action for the following question: $q$\\
---
\end{prompt}